\documentclass{article}
\usepackage{iclr2027_conference,times}
\usepackage{arxiv-preprint}

\usepackage{amsmath}
\usepackage{amssymb}
\usepackage{booktabs}
\usepackage{float}
\usepackage{graphicx}
\usepackage[hidelinks]{hyperref}
\usepackage{microtype}
\usepackage{multirow}
\usepackage{pgfplots}
\usepackage{tabularx}
\usepackage{tikz}
\usepackage{xurl}
\usepackage{xcolor}
\usepackage{enumitem}

\usetikzlibrary{arrows.meta,backgrounds,calc,fit,positioning}
\usepgfplotslibrary{groupplots}
\pgfplotsset{compat=1.18}
\hypersetup{hidelinks}

\definecolor{oiBlue}{HTML}{0072B2}
\definecolor{oiOrange}{HTML}{E69F00}
\definecolor{oiGreen}{HTML}{009E73}
\definecolor{oiVermillion}{HTML}{D55E00}
\definecolor{oiPurple}{HTML}{7A3E9D}
\definecolor{oiSky}{HTML}{56B4E9}
\definecolor{oiPink}{HTML}{CC79A7}

\newcommand{\rhae}{\textsc{RHAE}}
\newcommand{\arcagi}{ARC-AGI-3}
\newcommand{\dev}{\mathcal{D}_{\mathrm{dev}}}
\newcommand{\test}{\mathcal{D}_{\mathrm{test}}}

\title{Can LLMs Improve Their Own Harness?\\
A Controlled Study on ARC-AGI-3}

\author{Jens Lehmann$^{1,2}$ \qquad Sahar Vahdati$^{3,4}$\\[0.4em]
\normalfont $^1$Amazon\thanks{Work done outside of Amazon.}\\
\normalfont $^2$Technische Universit\"at Dresden\\
\normalfont $^3$Universit\"at Hannover\\
\parbox[t]{0.97\textwidth}{\normalfont $^4$TIB -- Leibniz Information Centre
for Science and Technology and University Library}\\[0.4em]
\normalfont\texttt{jens.lehmann@tu-dresden.de}\quad
\texttt{sahar.vahdati@tib.eu}}
\hypersetup{
  pdftitle={Can LLMs Improve Their Own Harness? A Controlled Study on ARC-AGI-3},
  pdfauthor={Jens Lehmann, Sahar Vahdati},
  pdfsubject={Autonomous harness construction and adaptation},
  pdfkeywords={large language models, self-improvement, agent harnesses, ARC-AGI-3}
}

\begin{document}
\maketitle

\begin{abstract}
A generally intelligent system acquires new skills efficiently without
requiring humans to redesign its software scaffold for every new task.
Yet language models rarely act alone.
A harness is a symbolic program that orchestrates calls to the model,
handling perception, memory, tool use, recovery, and inference allocation
on the model's behalf.

\arcagi{} is an interactive benchmark in which agents must explore unfamiliar environments, figure out their goals on the fly, and keep learning as they go. 
On it, human-steered systems already report ceiling scores across the public games, so we know effective designs exist.
We ask whether frontier models can build and improve such a harness on their own.
Through a fixed coding bootstrap, each model constructs a complete task harness, acts through that harness on five development games, and revises the entire codebase twice.
Every version is evaluated on twenty games that never influence adaptation.
We measure task performance with Relative Human Action Efficiency (\rhae{})
and record dollar-denominated inference cost.
Across 27 lineages, harness performance ranges from zero to near-ceiling scores.
Two revisions raise mean held-out \rhae{} by only 2.80 points
(95\% interval $[-1.44,7.35]$).
Development gains overstate held-out improvement, and
reductions in inference cost can come at the expense of task performance.
No revision retains its parent, yet a post hoc rule that selects among
existing snapshots by development score raises the held-out gain to 4.53
points ($[0.81,8.68]$), separating the generation of revisions from the
decision to deploy them.
Human-steered systems demonstrate further headroom.
These findings reveal a gap between autonomous harness construction and reliable adaptation under this two-revision protocol: even strong task performance does not ensure that subsequent revisions generalize or preserve existing capabilities.
\end{abstract}

\section{Introduction}
\label{sec:introduction}

Modern language models can write programs, call tools, and act in unfamiliar environments, but only as part of a larger agentic system. In such a system, a symbolic program orchestrates calls to the model, shaping what it observes, what it retains, which actions are available, how it recovers from failure, and how much inference it spends. 
This surrounding system is commonly termed a harness \citep{lee2026metaharness,lin2026agenticharness}. 
A system's behavior therefore depends on both its harness and its underlying model or models.

This matters for the study of general intelligence, which requires the
efficient acquisition of useful new skills in unfamiliar situations
\citep{chollet2019measure}. 
If each novel task requires a human engineer to inspect failures, redesign representations, add tools, and tune control flow, then what improves is not the system's own capacity to adapt but the engineer's understanding, transferred into the system by hand.
The agent's learning remains confined to a supplied mechanism whose improvement depends on the engineer. 
In contrast, humans routinely adapt to unfamiliar situations in everyday life.

AI systems can improve through changes to their underlying models or to the
harnesses through which those models act. These routes are complementary:
harness changes can provide access to new tools, maintain persistent memory,
and coordinate multiple model calls, capabilities that changes to model weights alone do not supply. 
The harness route therefore raises a central
question: 

\textbf{Can a model take on part of the engineering work that determines its own capabilities, using its execution experience to improve the software through which it acts?}

\arcagi{} aims to measure general intelligence through agents' adaptation to
novel games. Its main metric, Relative Human Action Efficiency (\rhae{}),
combines level completion with action efficiency relative to human baselines.
Harnesses developed under human steering already reach 100 \rhae{} on the
public games with GPT-5.6 Sol and Opus 5
\citep{lehmann2026tycho,han2026vista,chen2026avoarc}. 
By contrast, Sol scores 38.3 on the public set in a simple benchmark harness with retained reasoning and compaction enabled \citep{openai2026arcagi3settings}.
An off-the-shelf Claude Code baseline scores 42.83 with Opus 4.8/Fable 5
\citep{zeng2026schema}. %
Strong human-steered results make \arcagi{} a revealing test of autonomous
improvement: effective harness designs are known to exist for current models.
The benchmark lets us examine how reliably models can discover and improve
their own designs through experience in unfamiliar interactive environments.

Recent systems automate agent and harness design by searching over prompts,
skills, workflows, tools, memory, and source code
\citep{zelikman2024stop,hu2025adas,zhang2025aflow,lee2026metaharness,
lin2026agenticharness,zhang2026darwin}. 
Our replicated study examines how reliably revisions generalize across model conditions, evaluating every prescribed harness snapshot on held-out games and measuring gains, regressions, and cost.
No single study settles whether models can improve the software through
which they act. Our aim is one controlled, replicated measurement under an
explicit protocol, complementing studies that use other tasks, seeds, and
update rules.

We ask a direct question: \emph{Can a frontier reasoning model construct a
complete interaction harness, learn from its execution record, and improve
that same harness without human edits?} We measure the outcome of updating
with a fixed-lineage protocol. Through a fixed coding bootstrap, a model
starts from a controlled minimal repository, builds an \arcagi{} harness for
the same model profile, receives feedback from five development games,
and potentially revises the generated harness twice. 
We then evaluate the initial, revision-1, and revision-2 snapshots on twenty holdout games.

Our results expose a gap between autonomous harness engineering and reliable
self-improvement. Models autonomously produce diverse harnesses spanning
zero to near-ceiling performance. Yet their revisions do not
consistently turn development experience into held-out gains: mean
improvement on development games exceeds held-out improvement by 5.62
\rhae{} points, and cost reductions can accompany performance losses.
The updaters never retain a parent, although a simple development-based
choice among existing snapshots recovers part of the benefit.
These findings show why the capacity to build an agent and the capacity to
improve it need to be evaluated separately.
We make three contributions:
\begin{itemize}[nosep]
    \item We provide a general framework for autonomous \emph{harness
    adaptation} and define the harness adaptation gap: the difference between
    improvement on development tasks and improvement on held-out tasks.
    \item We introduce a replicated, all-snapshot protocol and conduct a
    controlled study across nine model--effort conditions: 27 independent
    harness lineages, 81 frozen snapshots, and 2,025 game evaluations,
    measuring task performance, deployment cost, and adaptation cost.
    \item We perform an outcome-blind, reliability-audited analysis of the
    observable capabilities in 54 revision episodes, connecting development
    feedback to code changes and relating those changes to performance on
    games withheld from revision.
\end{itemize}

\section{Harness Adaptation as a Capability}
\label{sec:framework}

Let a deployed agent be $S_t=(M,H_t)$, where $M$ is a pretrained model and
$H_t$ is an executable harness. For a task
$\tau$, the inner loop maps an interaction history to actions under a fixed
$H_t$. We write its expected utility over task distribution $\mathcal{D}$
as $J_{\mathcal{D}}(H_t;M)$, with $M$ held fixed as the harness changes.
The complete system is the unit of capability
measurement \citep{russell1995bounded,kapoor2025agents}.
An outer-loop updater receives development evidence
$E_t^{\mathrm{dev}}$ and produces
\[
 H_{t+1}=U(M,H_{\mathrm{boot}},H_t,E_t^{\mathrm{dev}}).
\]
The same model profile $M$ appears in both loops. In the outer loop, it
operates through a fixed general-purpose coding bootstrap
$H_{\mathrm{boot}}$, which supplies repository and development tools used to
construct and revise $H_t$. In the inner loop, it operates through the
task-specific harness $H_t$ to interact with the environment.
$H_{\mathrm{boot}}$ remains fixed while $H_t$ is the object of autonomous
adaptation.
Meta-learning methods such as MAML, RL$^2$, and AdA learn rapid adaptation
within a supplied neural architecture and training procedure
\citep{finn2017maml,duan2016rl2,bauer2023ada}. Here, pretrained weights remain
fixed while the model constructs and revises the executable task harness.

\begin{figure}[t]
\centering
\begin{tikzpicture}[
    box/.style={draw=black!55, rounded corners=1.5pt, align=center,
      font=\scriptsize,
      minimum height=7mm, inner xsep=3pt, fill=black!2},
    data/.style={box, fill=oiSky!12},
    update/.style={box, fill=oiOrange!15, text width=23mm},
    qualify/.style={data, text width=23mm},
    snapshot/.style={box, text width=15mm},
    eval/.style={box, fill=oiGreen!12},
    dev/.style={eval, text width=15mm},
    stage/.style={font=\scriptsize\bfseries, align=center},
    arr/.style={-{Latex[length=2mm]}, line width=.55pt},
    repair/.style={-{Latex[length=1.5mm]}, line width=.5pt,
      oiOrange!85!black},
    nofeed/.style={arr, dashed, black!50}
]
\node[stage] at (0,3.65) {Initial construction};
\node[stage] at (4.8,3.65) {Revision 1};
\node[stage] at (9.6,3.65) {Revision 2};

\node[update] (u0) at (0,2.45)
  {minimal repository $\rightarrow$ candidate\\same $M$, fixed $H_{\mathrm{boot}}$};
\node[qualify] (q0) at (0,1.15)
  {execution check\\2 development games};
\node[snapshot] (h0) at (0,0) {freeze\\$H_0$};
\node[dev] (d0) at (2.25,0)
  {inner loop\\5 development games};

\node[update] (u1) at (4.8,2.45)
  {revise from $E_0^{\mathrm{dev}}$\\same $M$, fixed $H_{\mathrm{boot}}$};
\node[qualify] (q1) at (4.8,1.15)
  {execution check\\2 development games};
\node[snapshot] (h1) at (4.8,0) {freeze\\$H_1$};
\node[dev] (d1) at (7.05,0)
  {inner loop\\5 development games};

\node[update] (u2) at (9.6,2.45)
  {revise from $E_1^{\mathrm{dev}}$\\same $M$, fixed $H_{\mathrm{boot}}$};
\node[qualify] (q2) at (9.6,1.15)
  {execution check\\2 development games};
\node[snapshot] (h2) at (9.6,0) {freeze\\$H_2$};
\node[dev] (d2) at (11.85,0)
  {development\\evaluation\\5 games};

\node[eval, text width=118mm] (held) at (5.9,-1.4)
  {all frozen snapshots $H_0,H_1,H_2$
   $\longrightarrow$ held-out evaluation (20 games each)
   $\longrightarrow$ outcomes only};

\begin{scope}[on background layer]
  \foreach \u in {u0,u1,u2} {
    \foreach \layer in {3,2,1} {
      \path let \p1 = ($(\u.north east)-(\u.south west)$) in
        node[box, fill=oiOrange!15, inner sep=0pt,
          minimum width=\x1, minimum height=\y1, rotate=2*\layer]
        at ([xshift=\layer*.65mm,yshift=\layer*.75mm]\u.center) {};
    }
  }
\end{scope}

\foreach \u/\q/\h in {u0/q0/h0,u1/q1/h1,u2/q2/h2} {
  \draw[arr] (\u) -- (\q);
  \draw[arr] (\q) -- node[left, font=\tiny] {pass} (\h);
  \draw[repair] (\q.east) to[out=0,in=0,looseness=1.15]
    node[right, font=\tiny] {fail}
    (\u.east);
}
\draw[arr] (h0) -- (d0);
\draw[arr] ([xshift=-2mm]d0.north east) |- (u1.west);
\node[font=\tiny, anchor=east] at (2.84,0.92) {$E_0^{\mathrm{dev}}$};
\draw[arr] (h1) -- (d1);
\draw[arr] ([xshift=-2mm]d1.north east) |- (u2.west);
\node[font=\tiny, anchor=east] at (7.64,0.92) {$E_1^{\mathrm{dev}}$};
\draw[arr] (h2) -- (d2);
\draw[nofeed] (h0.south) -- (h0.south |- held.north);
\draw[nofeed] (h1.south) -- (h1.south |- held.north);
\draw[nofeed] (h2.south) -- (h2.south |- held.north);
\end{tikzpicture}
\caption{\textbf{Controlled harness-adaptation workflow.} At each stage, the same model
profile uses the fixed coding bootstrap to propose a candidate.
The controller checks that it can execute and interact with two development
games. A failed candidate may be repaired; a passing candidate is frozen as
$H_t$. Five-game development evidence from $H_0$ and $H_1$ seeds
the next revision. $H_2$ runs on the same games for evaluation only.
All snapshots are evaluated on twenty held-out games after adaptation.}
\label{fig:loops}
\end{figure}
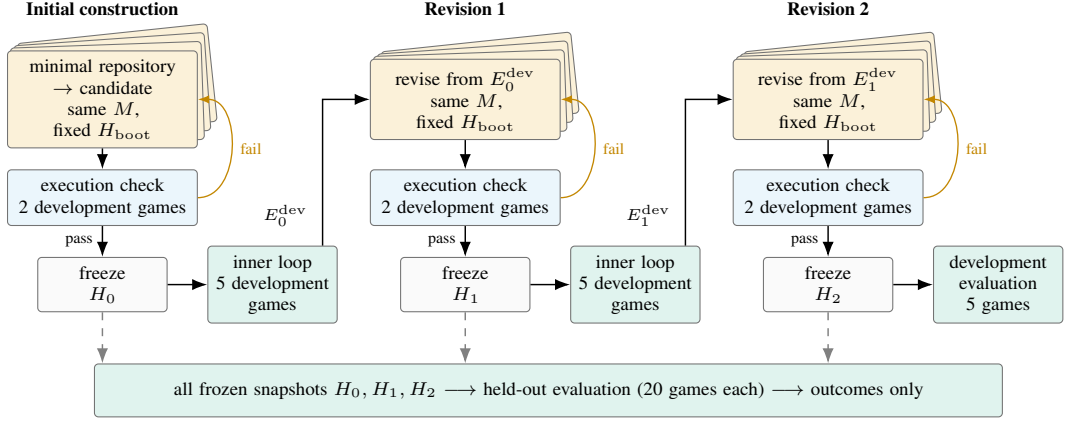

We separate four capabilities: (i) constructing an executable harness,
(ii) repairing interface and lifecycle failures, (iii) mapping execution
evidence to a concrete update, and (iv) selecting updates that improve
feedback-held-out tasks. The first three do not imply the fourth. A capable
updater must also
preserve working behavior and abstain when evidence is weak.

For revision $t$, define the harness adaptation gap
\[
G_t =
\left[J_{\dev}(H_t)-J_{\dev}(H_0)\right]
-
\left[J_{\test}(H_t)-J_{\test}(H_0)\right].
\]
A positive $G_t$ means that apparent adaptation on feedback-generating tasks
exceeds held-out benefit. Let $C_{\mathrm{run}}(H_t)$ denote the expected
execution cost per task and $C_{\mathrm{outer}}(H_{0:t})$ the cumulative cost
of constructing and revising the harness through stage $t$.
We evaluate performance and both costs jointly:
\[
 \left(J_{\test}(H_t),-C_{\mathrm{run}}(H_t),
 -C_{\mathrm{outer}}(H_{0:t})\right).
\]
This follows bounded-optimal and resource-rational accounts in treating
computation as a costly action \citep{russell1991metareasoning,
lieder2017strategy,lieder2020resource}. Our Pareto comparisons concern
held-out performance and per-game deployment inference cost; construction
and revision expenditure is reported separately.

In our \arcagi{} instantiation, task utility is a game's \rhae{} score,
and we estimate $J_{\dev}$ and $J_{\test}$ by mean \rhae{} over the
five development and twenty held-out games, respectively. We record
$C_{\mathrm{run}}$ and $C_{\mathrm{outer}}$ as dollar-denominated inference
costs for gameplay and harness development.

\section{Related Work}
\label{sec:related-work}

\textbf{LLM self-refinement and program optimization.}
Self-Refine, Reflexion, and CRITIC revise responses or task policies through
self-feedback, retained verbal feedback, or tool-grounded critique
\citep{madaan2023selfrefine,shinn2023reflexion,gou2024critic}. Prompt and
program optimizers learn reusable components from development data. OPRO and
PromptBreeder search instruction space. DSPy, MIPRO, TextGrad, and GEPA
optimize modules inside declared language-model programs
\citep{yang2024opro,fernando2024promptbreeder,khattab2024dspy,
opsahlong2024mipro,yuksekgonul2025textgrad,agrawal2026gepa}. These methods
normally retain a human-specified artifact schema or computation graph.

\textbf{Automated agent and harness design.}
Prior work automates optimizer construction, code-defined agent design,
and workflow search
\citep{zelikman2024stop,hu2025adas,zhang2025aflow}, and extends autonomous
revision to runtime logic and complete coding-agent repositories
\citep{yin2025godelagent,robeyns2025sica,zhang2026darwin}. Meta-Harness and
Agentic Harness Engineering optimize executable harness source from scores and
traces \citep{lee2026metaharness,lin2026agenticharness}. HarnessCompass and
RRSI study how such changes generalize beyond the tasks used for evolution
\citep{zhang2026harnesscompass,xia2026rrsi}. We study autonomous harness
engineering as a capability of frontier models: how reliably can they
construct the software through which they act and improve it from their own
execution evidence? 
Our replicated comparison follows each independently
constructed harness and both prescribed revisions, measuring initial
capability alongside subsequent gains and regressions on games excluded
from adaptation.

\textbf{Evaluating autonomous improvement.}
SEAGym reports a schedule that completes 39 of 40 harness updates while
test success falls by 14.5 percentage points \citep{zheng2026seagym}.
On Terminal-Bench, validation-selected evolution yields little mean
held-out gain, whereas on \arcagi{} a harness that adapts within each
evaluation game improves substantially \citep{wang2026rethinking};
HarnessOpt-Bench finds that the
best validation score seen during search is usually optimistic relative to
the nominated harness's test score \citep{ursekar2026harnessopt}.
RRSI obtains transfer over 20--40 rounds with explicit proposal and
acceptance regularization \citep{xia2026rrsi}.
Our construction-and-revision study extends this evidence to independently
generated interactive-game harnesses, asking whether revisions based on some
games transfer to others, and jointly tracking performance on
development and held-out games and inference cost.

\textbf{Harness design for ARC-AGI-3.}
EWM, OPINE-World, PRO-LONG, and Tycho demonstrate strong autonomous game
solving within human-steered harnesses
\citep{rodionov2026executable,courtis2026opine,fox2026prolong,
lehmann2026tycho}. Workspace Optimization lets agents revise executable
world models and strategies during play while the reusable harness remains
fixed \citep{sarafian2026workspace}. Our study shifts harness design and
revision to the model: it constructs the reusable task harness and revises
it from its own execution evidence, without human edits or guidance between
stages. We evaluate whether those revisions improve performance on games
excluded from adaptation. Appendix~\ref{app:related} compares the protocol
designs.

\section{Experimental Design}
\label{sec:design}

\subsection{ARC-AGI-3: An Interactive Reasoning Challenge}

Unlike the static input--output transformations in earlier ARC tasks,
\arcagi{} presents an agent with sequential, unfamiliar grid worlds.
At each turn, the agent observes one or more $64\times64$ color grids,
together with the current game state, progress across levels, and available
actions. Depending on the game, the interface offers directional inputs, a
game-specific input, coordinate clicks, or undo. The agent is not told what
the visible objects represent, how these inputs affect them, or what
completes a level. It must discover the game's rules through exploration and
reasoning.

Each game contains multiple levels. Relative Human Action Efficiency
(\rhae{}) assigns credit only to completed levels and discounts solutions that
use more environment actions than hidden human baselines. Later levels receive
greater weight. We also record actions, completed levels, terminal states, and
dollar-denominated model inference cost.

\begin{figure}[t]
\centering
\setlength{\fboxsep}{1pt}
\fbox{\includegraphics[width=.29\linewidth]{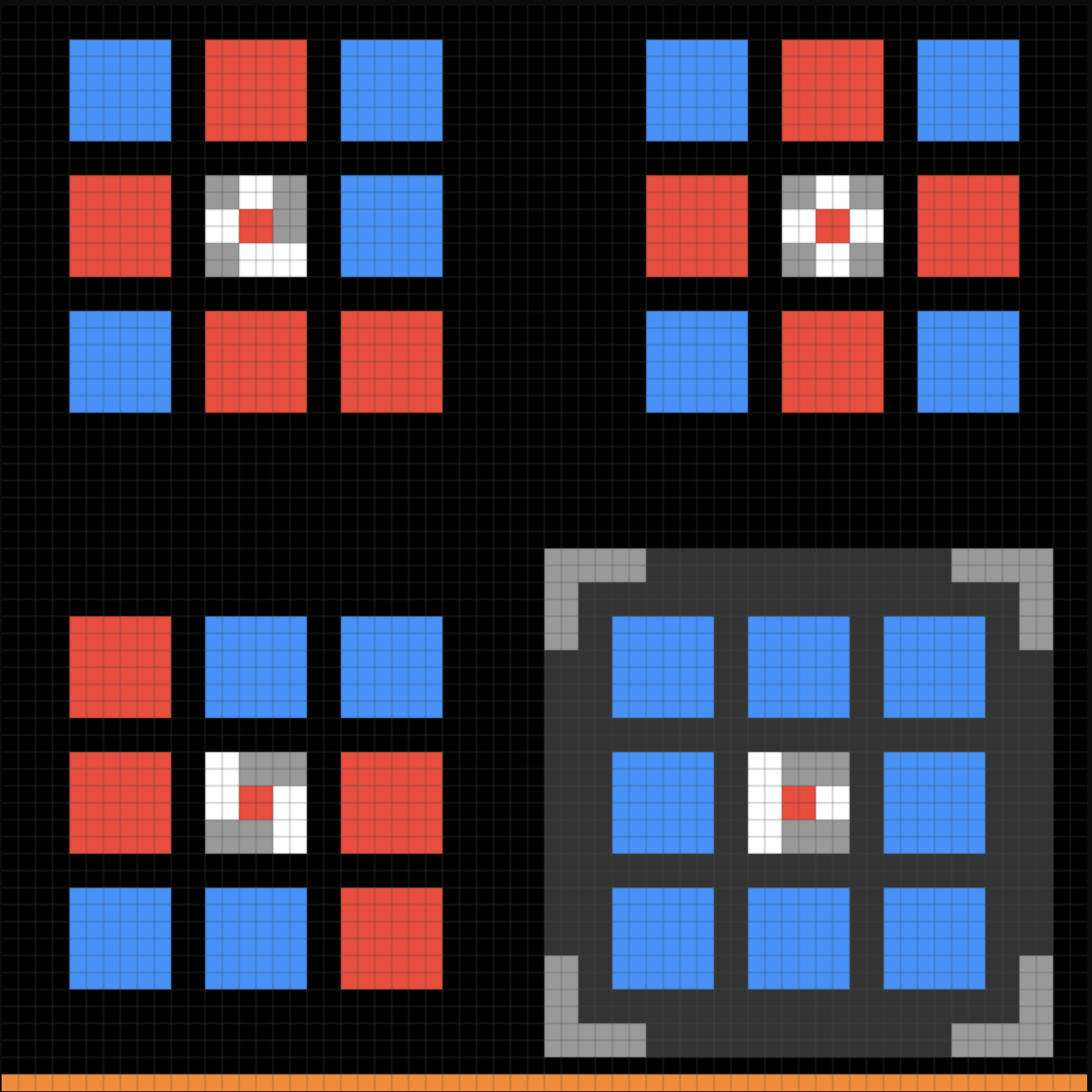}}
\hspace{5mm}
\fbox{\includegraphics[width=.29\linewidth]{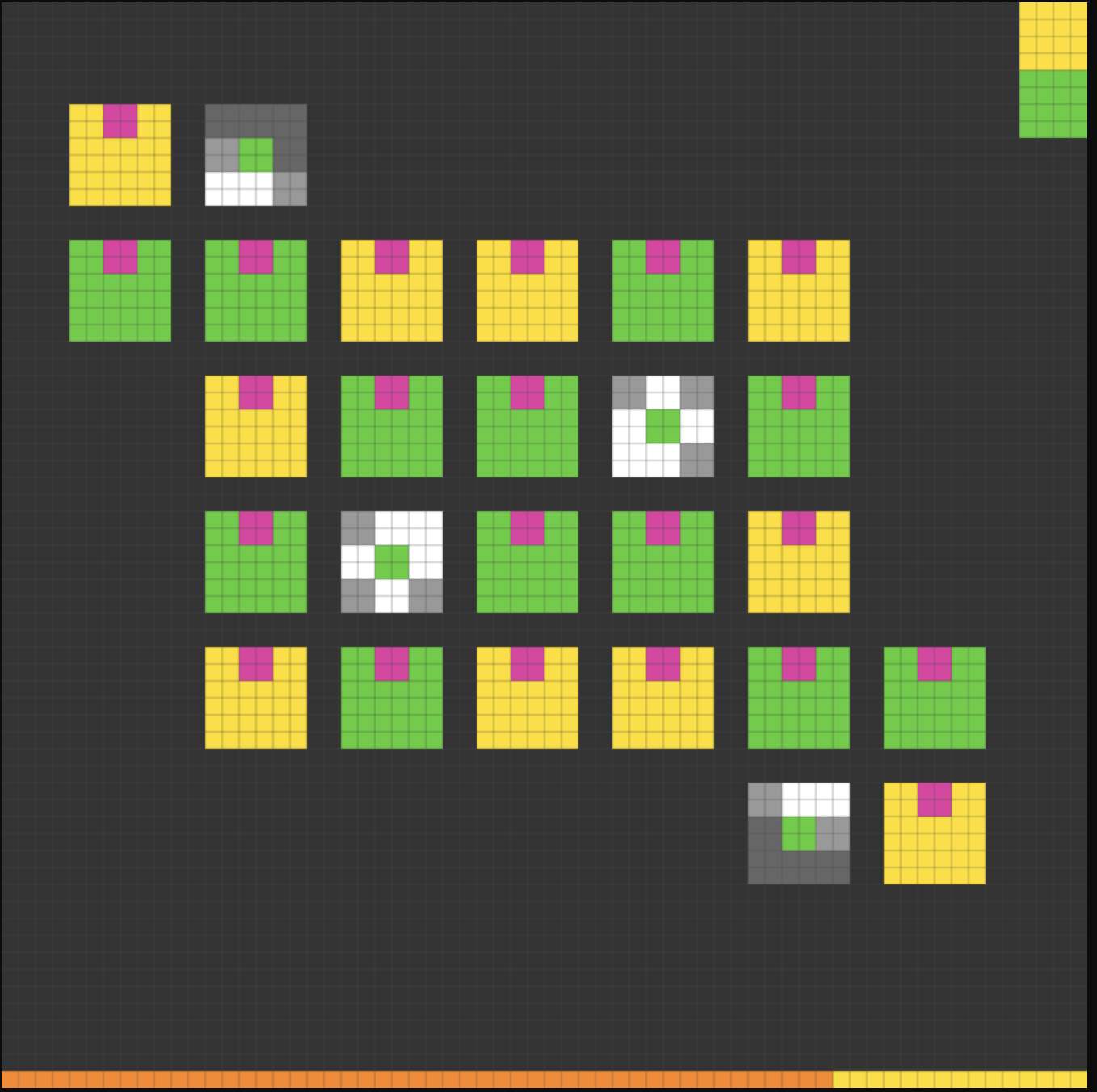}}
\caption{\textbf{Two observations from FT09 (Level 1 on the left, and Level 7 on the right), a public \arcagi{} game.}
The agent must infer that the small patterned tiles specify which neighboring
tiles must match or differ from a reference color. Clicking cycles tile
colors, which need to be matched to target patterns. The later levels introduce controls that change several tiles together.
These rules must be discovered through interaction.}
\label{fig:arc-game}
\end{figure}

Figure~\ref{fig:arc-game} shows two states from one such game, FT09 \footnote{\url{https://arcprize.org/tasks/ft09}}.
A harness can present these states to the runtime model as raw images,
symbolic grids, object summaries, transition diffs, or combinations of them.
It also controls context and memory, provides tools, organizes planning and
reasoning, and allocates model calls. It can request one action or a sequence,
use code to execute a discovered strategy, and reuse computations to reduce
inference cost. These design choices are not specified by the benchmark.

\subsection{Autonomous construction and revision}

\paragraph{Two harnesses, one model.}
Each experimental condition pairs a model with a reasoning-effort setting
and uses that profile for both harness development and gameplay.
During construction and revision, the model operates through OpenCode, a
general-purpose coding harness in the same category as Claude Code and Codex.
OpenCode provides tools to inspect and edit files and run commands. It
remains fixed throughout the campaign. During gameplay, the model is called
through the generated task harness, the artifact being autonomously
constructed and improved.

Each run receives a sandboxed workspace with Python, preinstalled scientific
libraries, and shell tools.
The repository includes documentation for interacting with the game, a client
for calling the assigned model, and model and pricing specifications.
The model must create \texttt{run.sh}, which launches its solver on a specified
game without human input. The solver architecture is otherwise unspecified.
The harness controls how much work is performed by deterministic code and
when the model is consulted.
The build model may create and revise solver code and run local checks,
including mock-based tests.

Coding sessions do not expose the evaluator's live game or runtime-model
gateways. End-to-end trials occur through controller-run smoke
qualification on two declared development games. If a candidate fails, the
same model receives its failure report and execution logs and may repair it
for at most ten cycles. Qualification requires the harness to run and interact
with the game within fixed action and level limits, regardless of its
\rhae{} score. Task performance is measured in the subsequent fixed-snapshot
evaluations.

The qualified initial system runs on five development games. Revision~1
receives a fixed bundle containing its own source, traces, scores, actions, and
inference costs. It may change any generated file or preserve the parent. The
same procedure produces revision~2. No human edits or natural-language
guidance are introduced between stages.
Resource limits and infrastructure-recovery rules are specified in
Appendix~\ref{app:protocol}.

To assess the outcomes of construction and revision, we compare the resulting
harness versions under a common evaluation protocol.

\begin{list}{\textbullet}{%
  \setlength{\leftmargin}{1.2em}%
  \setlength{\labelwidth}{.7em}%
  \setlength{\labelsep}{.5em}%
  \setlength{\topsep}{0pt}%
  \setlength{\partopsep}{0pt}%
  \setlength{\itemsep}{2pt}%
  \setlength{\parsep}{0pt}%
}
\item \textbf{Held-out evaluation.} Every frozen harness version is evaluated on the five development games and
twenty held-out public games. The held-out games provide no feedback for
harness construction, revision, or selection. Our primary comparison is the
change in mean held-out \rhae{} from the initial harness to revision~2 among
runs completing all three stages. Construction failures are reported
separately over all planned runs. We validate recorded gameplay by replaying
the same actions locally and through the official ARC service, checking
agreement in scores and game progress (Appendix~\ref{app:integrity}).

\item \textbf{Conditions and uncertainty.}
The study covers nine model--effort conditions across seven model families,
with three independent constructions per condition
(Table~\ref{tab:conditions}). All 27 runs complete construction and both
revisions. All conditions follow the same
construction-and-revision protocol. We estimate confidence intervals by
bootstrapping whole lineages, preserving paired comparisons across versions.
These intervals describe uncertainty for the observed study pool on the fixed
game split. Comparisons between model--effort conditions are descriptive.
\end{list}

\section{Quantitative Results}
\label{sec:results}

Held-out gains are small on average and vary substantially across lineages.
Revision~2 raises mean development \rhae{} by 8.42 points and held-out
\rhae{} by 2.80 points. The held-out 95\% bootstrap interval is
$[-1.44,7.35]$, compatible with no average benefit or a useful positive gain;
with 27 lineages, the design detects mean changes of about 6.5 points with
80\% power (Table~\ref{tab:power}). The interval's upper end is small
against the headroom that human-steered harnesses demonstrate for some of
the same models (Section~\ref{sec:discussion}).
At a descriptive five-point threshold, 9 lineages gain, 3 lose, and 15 change
by less; strict signs give 15 gains and 12 losses ($p=.70$, two-sided sign
test; Appendix~\ref{app:quantitative}). These are observed suite outcomes,
not estimates of how often revisions reduce expected performance.

Each snapshot is evaluated once, so we measured execution variability
separately. Ten evaluations of one frozen final harness on the same twenty
held-out games yield a sample SD of 3.99 \rhae{}.
All ten scores remain above this lineage's initial-harness score, showing
that its large observed improvement persists across repeated evaluations.
The observed variation also shows why score
changes of only a few points require cautious interpretation
(Appendix~\ref{app:fixed-harness-variance}).
Paired repeats of three prespecified lineages reproduce the direction of both
a large gain ($+30.82$, $[9.97,51.67]$) and a large regression ($-16.57$,
$[-28.16,-4.99]$, an interval that reaches the five-point band), while a
near-zero change remains unresolved (Table~\ref{tab:paired-repeats}).

The pooled result also hides differences between games. Averaging across runs,
held-out performance improves on 13 games and declines on seven. The largest
mean gain is 20.44 \rhae{} points and the largest decline is 15.03.
Aggregate improvement can therefore coexist with substantial regressions on
particular games (Appendix~\ref{app:quantitative}).

\begin{figure}[t]
\centering
\includegraphics[width=\linewidth]{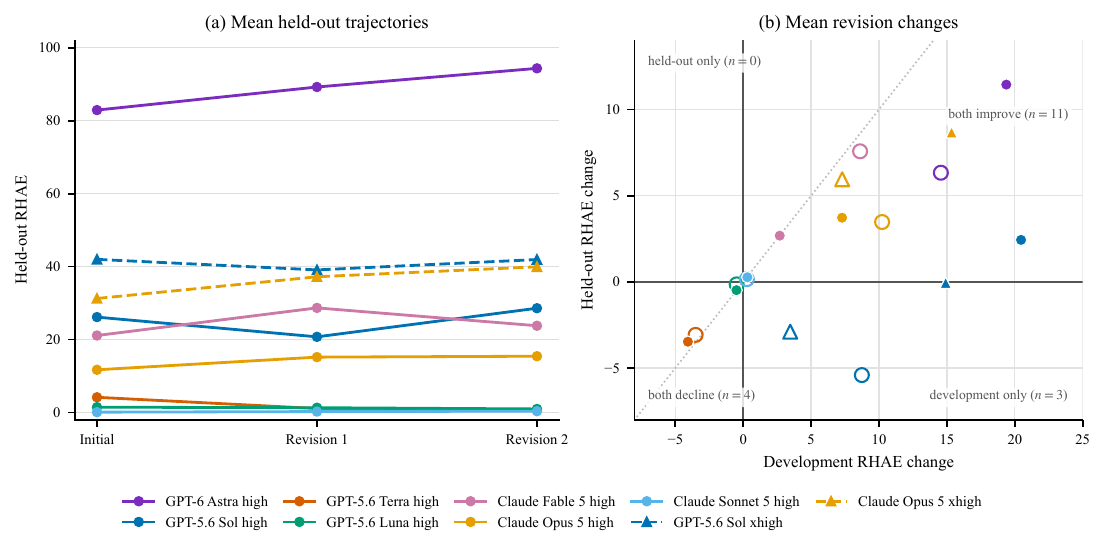}
\caption{\textbf{Construction strength varies, and development gains transfer
unreliably.} (a) Held-out condition means on a common scale.
(b) Each condition has two points, averaging its three runs:
initial-to-revision-1 (open) and initial-to-revision-2 (filled). Labels
count strict-sign quadrants. Points below zero regress on held-out games.
Triangles denote higher-effort conditions.}
\label{fig:main-results}
\end{figure}

Development gains overstate transfer.
The mean adaptation gap $G_2$ is 5.62 points (95\% bootstrap interval
$[1.78,9.77]$). Figure~\ref{fig:main-results}b shows several revisions that
improve the five feedback-generating games while reducing held-out
performance. Development improvement therefore overstates the average
held-out gain under this update procedure.
Before revision the two splits score alike (26.13 development and 24.54
held-out \rhae{}); the gap opens with revision.
Its point estimate stays positive in every sensitivity analysis, but its size
is uncertain: it ranges from 4.07 to 6.40 when one condition is omitted and
from 3.02 to 7.84 when one development game is omitted from the development
summary, and it is 4.21 ($[-0.05,8.87]$) when games affected by execution
recovery are excluded (Appendix~\ref{app:quantitative},
Table~\ref{tab:recovery-sens}). Without the three near-floor conditions, the
held-out gain is 4.82 ($[-1.44,11.38]$) and the gap 8.52 ($[3.16,14.11]$;
Table~\ref{tab:floor}).

\textbf{Choosing what to deploy recovers part of the benefit.}
A post hoc rule selecting the snapshot with highest development \rhae{}
yields a mean held-out gain of 4.53 points over the initial harness
(95\% interval $[0.81,8.68]$), compared with 2.80 for always deploying $H_2$.
It retains $H_0$ in nine lineages and reduces observed regressions from 12 to
5, at a mean deployment cost of \$9.15 rather than \$7.57 per game.
The rule was specified after held-out results were available. Its gain over
$H_0$ is stable when any one development game is omitted from selection
($+3.95$ to $+4.16$), whereas its advantage over always deploying $H_2$ is
smaller and less certain (Table~\ref{tab:selection-loo}); the rule remains
to be evaluated prospectively.

Lower cost does not indicate improvement.
For each immutable snapshot, solving cost is the mean logical inference cost
over the 20 held-out games, including replayed responses. Cost per game
falls in 18 of the 27 lineages. At the five-point threshold, five of these 18
also gain held-out \rhae{}, three lose, and ten change by less. By strict
sign, nine revision-2 descendants improve both held-out \rhae{} and cost per
game, six gain \rhae{} at higher cost, nine become cheaper while losing
\rhae{}, and three worsen on both dimensions; the paired repeats support no
strict score--cost classification, so these are observed outcomes
(Table~\ref{tab:paired-repeats}). Mean cost falls by \$2.72 per game, but
the median change is only $-\$0.54$. The lower average expenditure thus
combines revisions that achieve more at lower cost and revisions that
spend less while also solving less effectively.

\begin{figure}[t]
\centering
\includegraphics[width=\linewidth]{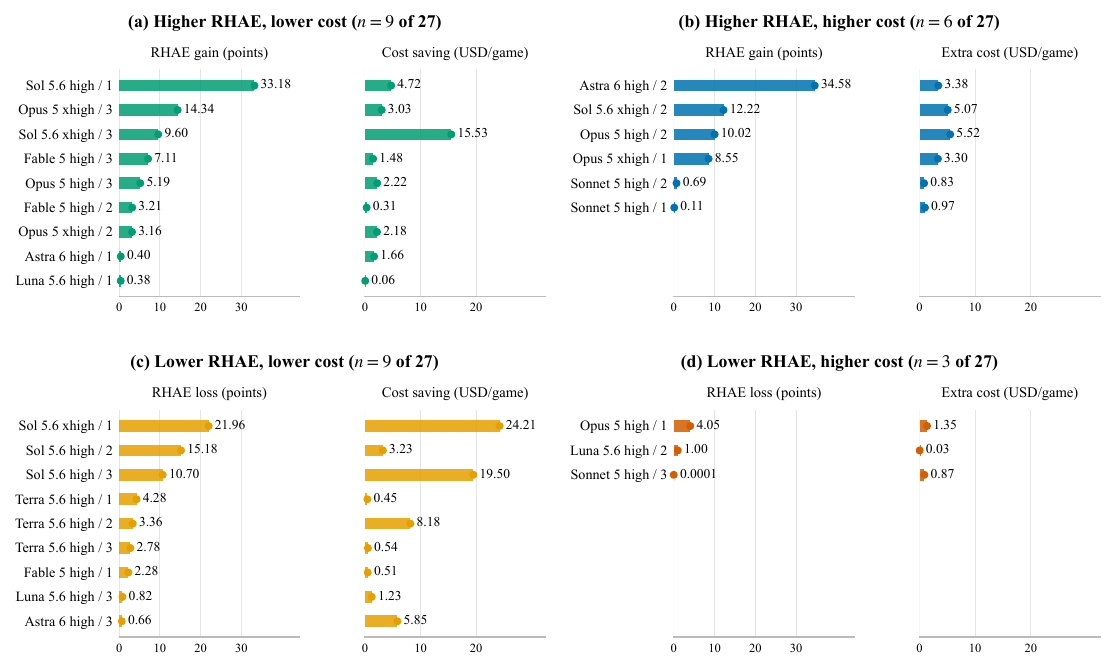}
\caption{\textbf{Separate outcomes reveal which harnesses improve, save cost,
or trade one for the other.} Each row identifies a model--effort condition
and run. Bars show magnitudes of observed $H_2-H_0$ changes in mean held-out
\rhae{} and inference cost per game; headers specify the direction.
Each metric uses a common linear scale across panels. Categories use strict
signs of unrounded changes; Table~\ref{tab:threshold-sensitivity} gives
counts under practical thresholds.}
\label{fig:cost-outcomes}
\end{figure}

The experimental program corresponds to at least \$18,129.27 under the frozen
publication price schedule, including construction, revision, repair,
qualification, and fresh selected-game inference.
Appendix~\ref{app:quantitative} provides the cost breakdown, complete
replicate-level trajectories (Table~\ref{tab:replicate-score-cost}), and
checkpoint and game-sensitivity analyses.

\section{What Models Build and How They Revise}
\label{sec:capabilities}

\subsection{What harness designs emerge, and how do they change?}

The generated systems span markedly different interaction strategies.
Compact direct model controllers coexist with heuristic-first
systems, strategist--executor hybrids, persistent manuals, explicit world
models, replay systems, deterministic planners, and inference-budget managers.
Architectures also vary across replicates of the same condition.

Across all 27 lineages, we audit all 54
parent-to-descendant transitions.
This post hoc descriptive audit labels a revision as
\emph{Preserve} when it repairs or tunes the current policy, \emph{Extend} when
it adds a capability inside the same decision loop, \emph{Reorient} when it
materially changes control priority, context continuity, replay, or
action/inference scheduling, and \emph{Replace} only when it changes the
harness family. These labels describe observable source and runtime structure;
they are separate from the outcome-blind capability coding below.

Strategy change is common. Twenty-two of 27 first revisions reorient the harness,
four extend it, and one primarily preserves or restores the incumbent.
Second revisions include 12 reorientations, 12 extensions, and three
preserves. No lineage replaces its harness family; revisions repeatedly
reorganize memory, perception, batching, replay, and resource allocation.
Appendix~\ref{app:case-studies} reports every replicate and
Figure~\ref{fig:harness-evolution} summarizes size and code churn.

One Sol-high revision addressed decision-limit failures and ineffective
memory use by changing action budgeting, memory management, and
observations. Held-out \rhae{} rose from 29.24 to 53.56 at essentially unchanged
inference cost (Appendix~\ref{app:case-studies}).
In a Sol-xhigh lineage, revisions replaced response chaining with independent
calls and bounded memory. Held-out \rhae{} fell from 43.95 to 21.99 as
inference cost fell from \$31.39 to \$7.18 per game
(Appendix~\ref{app:case-studies}).

\subsection{Which revision behaviors accompany held-out improvement?}
\label{sec:revision-behavior}

We examine how models respond to execution feedback and whether their
revision behaviors accompany held-out gains.

\paragraph{Corpus and codebook.}
The coding corpus contains all 54 parent-to-descendant revision episodes
across the nine conditions, comprising 70 updater sessions.
Packets preserve the development
feedback available to the updater, compact chronological messages and tool
events, the exact selected-system diff, and proximal test and smoke results.
They mask model and condition identity, trajectory identifiers, future stages,
and every downstream development and held-out outcome. The codebook covers
evidence use, planned and enacted changes, verification, and post-result
adaptation (Appendix~\ref{app:codebook}).

\paragraph{Blinding and reliability.}
Each outcome-blind packet was independently coded twice using GPT-5.6 Sol
xhigh under the same frozen codebook. Consensus labels were frozen
before linkage to downstream outcomes.
We report unadjudicated exact agreement, quadratic-weighted Cohen's $\kappa$
\citep{cohen1968weighted}, and Krippendorff's $\alpha$
\citep{krippendorff2004reliability}. Scope
($\kappa=0.96$), post-result adaptation ($0.84$), and
planning ($0.70$) pass the exploratory
reliability threshold. Updater verification is uniformly top-coded with exact
agreement, so its chance-corrected reliability is undefined.
The planning interval is absent from 27 packets, with
missingness concentrated in several conditions and larger edits; we therefore
exclude it from outcome interpretation. Diagnostic probing ($0.66$),
diagnosis ($0.35$), incumbent protection ($0.47$), and resource metareasoning
($0.29$) fall below the preset $0.67$ reliability threshold. Visible
narration is triangulated against tools, code, tests, and runtime behavior
because verbalized reasoning may be unfaithful
\citep{turpin2023unfaithful,lanham2023faithfulness,
chen2025reasoningfaithfulness}.

\paragraph{Do broader edits accompany larger gains?}
All 54 revision episodes include inspection of item-specific development evidence, all final diffs
target at least one eligible evidence-backed issue, and every selected
descendant passes its proximal operational checks. Completing an
evidence-guided revision does not ensure a better agent: all episodes satisfy
these process checks, while held-out outcomes vary. Broad rewrites average $+3.75$ immediate
held-out change versus $-0.20$ for contained changes, with a wide interval
that includes zero; a deterministic rule-based re-derivation of scope agrees
with the consensus labels in 51 of 54 episodes and reproduces this contrast
(Table~\ref{tab:coding-validation}).
Appendix~\ref{app:codebook} reports the complete exploratory association
table.

\paragraph{How do reactive revision episodes differ?}
In 36 episodes, the updater links the final result to its change and hands off.
Their mean immediate held-out change is $+2.48$. In 18 episodes, a post-edit
result triggers another edit, revert, test change, scope change, or
criterion-based stop. Their mean is $-0.76$. The $-3.24$-point contrast
is uncertain: its interval includes zero. Episodes also differ in parent
performance, repair effort, and the changes attempted. The comparison
therefore establishes neither a dependable advantage nor a harmful causal
effect of further repair. A rule-based re-derivation of this code agrees
in 46 of 54 episodes, and under it the contrast disappears ($+0.48$), so this
association depends on how post-result adaptation is operationalized
(Table~\ref{tab:coding-validation}).

\section{Discussion}
\label{sec:discussion}

\paragraph{Construction strength does not ensure consistent improvement.}
Across nine conditions, the study establishes substantial autonomous
harness-building capability while exposing a gap between development gains
and held-out benefit. The pooled interval remains compatible with useful
average improvement, and several lineages achieve large gains. Reliability
here concerns consistent held-out benefit and capability preservation;
observed regressions do not establish each revision's effect on expected
performance. Evidence inspection, targeted edits, and operational validation
are common, but their exploratory associations cannot establish why revisions
succeed or fail. Strong initial harnesses can lose observed performance while
weaker constructions improve substantially.

\paragraph{A gap to human-steered harnesses remains.}
At the same Opus 5 xhigh setting, Tycho reaches 100 \rhae{} on every public
game, whereas our revision-2 harnesses average 39.95 on twenty of those games
\citep{lehmann2026tycho}. Both studies designate the same five development
games, although Tycho also uses the full public set for policy selection.
Tycho reports 100 with Sol, while VISTA and AVO report 100 with
Opus 5 \citep{han2026vista,chen2026avoarc}. Retrodict reports 99.86 with
Sol at maximum effort \citep{brown2026retrodict}.
These results establish substantial headroom for autonomous harness
engineering. The development and selection procedures differ. So does what
the model sees during play: the reference systems keep the complete
interaction record available (a typed history in Tycho, every frame in
VISTA, a replayable log in Retrodict), whereas our Sol harnesses use
stateless calls or short chains with bounded notes and our Opus harnesses
rely mainly on build-time code, consulting the model 14--18 times per game.
Tycho's reported totals imply roughly 600--1,060 model calls per game, but
spend alone does not account for the gap: Retrodict reports about \$26 per
game, less than two of our initial Sol-xhigh harnesses spent. Model strength
matters as well: GPT-6 Astra, whose self-built harnesses reach 92.6--97.6,
scores 99.9 on the official leaderboard with a simple harness that preserves
reasoning state \citep{arcprize2026astra}. The comparison therefore does not
isolate an effect of human steering.
Appendix~\ref{app:human-designed} gives the comparison scope.

\paragraph{Abstention is part of adaptation.}
Although retaining the parent is permitted, every scored trajectory changes
its implementation: each of the 54 revisions adds or removes at least 84
lines of code (median 449), including the four labeled \emph{Preserve}.
Reliable self-improvement includes recognizing when
the evidence favors keeping the current harness. The evidence for that
decision is often available: selecting among existing snapshots by
development score retains $H_0$ in nine lineages and raises the held-out
gain from 2.80 to 4.53. Generating a revision and deciding whether to deploy
it are therefore distinct parts of autonomous improvement, and the updaters
we study exercise only the first.
Candidate revisions should
be compared with a preserved baseline and rejected or rolled back when they
do not justify replacement. RRSI's reported gains with edit budgets and
noise- and cost-aware acceptance rules illustrate one way to make such
decisions explicit \citep{xia2026rrsi}.

\paragraph{Implications for autonomous improvement.}
Harness adaptation complements model training by changing the software
through which fixed model weights act
\citep{lee2026metaharness,lin2026agenticharness}.
For AI risk analysis, current task performance is only part of the picture:
a system may also be able to improve the software that determines its future
performance \citep{wijk2025rebench}. Model and harness improvement could
reinforce each other if better harnesses support AI research and better
models become more capable harness developers \citep{chen2026recursivesurvey}.
Evaluating this route requires tracking whether revisions generalize,
preserve prior strengths, and justify their cost.
Our experiment holds both the model and coding bootstrap fixed.
Claims about sustained capability acceleration or its risks require longer
runs, broader tasks, and updaters that can themselves change
\citep{wijk2025rebench,chen2026recursivesurvey}.

\section{Limitations}
\label{sec:limitations}

We use one benchmark, one set of feedback games, and two revisions. The
results therefore leave open how harness adaptation behaves on other tasks
or over longer improvement sequences. Because the games are public, some
models may have encountered them during training.

Three runs per condition leave uncertainty about small differences between
models, and 27 lineages detect only mean changes of several points.
Gameplay can also vary even with an unchanged harness, as our
ten-run diagnostic shows. We did not repeat every evaluation, so individual
gains or regressions may partly reflect this variation. These limits follow
from cost: at the recorded prices, a further revision stage would cost about
\$5,200 and ten evaluations of every snapshot about \$145,000
(Appendix~\ref{app:discussion}).
Because the same model builds the harness and plays through it, near-floor
results conflate construction and gameplay. In a crossed evaluation,
Sol-built harnesses do not lift Terra or Luna as players, so gameplay
limits their scores, and Sonnet-built harnesses stay near the floor when Sol
plays them, so harness quality limits Sonnet's (Table~\ref{tab:crossplay}). The
reverse cells, such as a Terra-built harness played by Sol, were not run. A
direct contamination test would require
non-public games.

Inference expenditure varies across harnesses and models. Reporting score
and cost together makes this visible, but does not separate model capability,
harness design, and spending into independent effects.

The revision-behavior annotations in Section~\ref{sec:revision-behavior}
were produced by two LLM annotation passes. A rule-based re-derivation of the
two interpreted codes agrees closely for scope and moderately for post-result
adaptation; agreement with independent human annotations was not evaluated. The coders use summarized execution records,
which can lose context or event order. We exclude unreliable or inadequately
observed labels. Episodes differ in their starting harness and the problems
they attempt to fix, so an association cannot show that a particular revision
behavior caused the outcome.

\section{Conclusion}
\label{sec:conclusion}

Harness adaptation offers a route to self-improving AI in which a system
develops the software through which it acts. Our \arcagi{} study demonstrates
substantial autonomous harness-building capability, but a less dependable
ability to convert execution experience into held-out gains. Development
progress can overstate improvement, and lower cost does not indicate it: of
18 cheaper final harnesses, only five gain more than five held-out points. A
simple development-based choice among existing snapshots, specified post hoc,
recovers part of the benefit, which suggests that deciding whether to deploy
a revision deserves as much attention as producing it. Our evidence is
strongest for the specified update-and-deploy protocol and speaks less
directly to autonomous improvement in general. Reported human-steered systems show that stronger
harnesses are possible for some of the same models. Strong performance from a
model--harness system should be distinguished from its ability to autonomously
improve the software through which it acts. Our results make this distinction
concrete: with model weights fixed, development gains and lower inference
costs can coexist with held-out regressions, making transfer and preservation
of existing capabilities essential tests of harness adaptation.
The central challenge is
to make improvements accumulate: extending what the system can do across
tasks while preserving what it already does well.
\label{main-text-end}

\clearpage
\section*{AI Use Statement}
Language-model assistants supported experimental-infrastructure
implementation, code and methodological review, literature search, interpretation checks, manuscript editing, and the
outcome-blind qualitative coding described above.
The authors had the initial idea, selected the protocol and data, verified the analyses and claims, authored the final text, and take responsibility for the work.

\section*{Reproducibility Statement}
Appendices~\ref{app:protocol} and \ref{app:prompt} specify the protocol and
starting prompt; Appendix~\ref{app:related} compares related systems.
Appendices~\ref{app:quantitative}--\ref{app:codebook} provide supplementary
results, strategy audits, and the coding method. The
 supplementary{} artifact contains
the seed repository, frozen prompts, profiles, and splits, the scoring and
feedback-curation code, all selected source snapshots, analysis tables,
deterministic reproduction scripts, and a checksum-backed manifest (Appendix~\ref{app:artifact}).

\bibliography{references}
\bibliographystyle{iclr2027_conference}

\appendix

\section{Full Experimental Protocol}
\label{app:protocol}

The nine model--effort conditions each have three independent constructions
and follow the same construction-and-revision protocol
(Table~\ref{tab:conditions}). All conditions enter the pooled analysis.

\paragraph{Repository boundary.}
The initial repository provides the task contract, local engine client,
controlled model client, pricing documentation, and the entry-point contract.
It does not provide a solver architecture, game-specific rules, worked
examples, or a reusable task policy. Generated systems have no internet
access, no cross-game state, and no access to other game instances.
The fixed coding bootstrap is OpenCode; the artifact records its binary and
adapter fingerprints. Codex is used separately for outcome-blind annotation.

\paragraph{Development split.}
The five development games, \texttt{tr87}, \texttt{vc33}, \texttt{r11l},
\texttt{bp35}, and \texttt{ft09}, are those designated at random by Tycho
\citep{lehmann2026tycho}; the split was fixed before the campaign.
The other 20 public games were fixed as feedback-held-out. The two smoke
games were chosen from the development split to exercise click- and
keyboard-based interaction.

\paragraph{Stage lifecycle.}
Each stage has one selected candidate. Controller-owned smoke qualification
checks interface validity and bounded interaction on two declared development
games. Each game permits 50 charged actions and stops after two completed
levels or a win. Passage requires a charged action, a clean exit, and a
registered stopping boundary on each game; no score threshold is used.
Failed candidates may be repaired by the same condition for
at most ten cycles. The selected initial candidate runs the five-game
development evaluation. Its frozen source and feedback bundle seed revision~1.
The same procedure seeds revision~2. All attempts, including failed repair and
report-finalization sessions, remain in the audit ledger.

\paragraph{Hosted replay.}
Generated systems interact only with the local engine. The controller freezes
complete action--observation traces, replays them locally, then submits them to
a short-lived hosted scorecard. Acceptance requires score agreement and
matching compared observations, with one explicitly logged visual-noise
exception described in Appendix~\ref{app:integrity}. Hosted service behavior
therefore cannot influence the generated policy.

\paragraph{Construction replacement policy.}
A construction may be replaced only for controller, provider-transport, or
hosted service failures that occur outside model-produced behavior. Syntax errors,
invalid protocol use, exhausted repair budgets, and failure to produce a
qualified harness remain model or harness outcomes. Every planned construction
in the retained study cohort produced a qualified initial system.

\begin{table}[H]
\caption{Study conditions. Builder and runtime profiles are
matched within each row. Context is the advertised window; max output is the
configured per-response gameplay cap. Coding caps are given below. Exact
model- and pricing-profile hashes are retained in run protocols.}
\label{tab:conditions}
\centering
\small
\begin{tabular}{lrrr}
\toprule
Model profile & Reasoning effort & Context & Max output \\
\midrule
GPT-6 Astra & high & 1.05M & 64K \\
GPT-5.6 Sol & high & 1.05M & 120K \\
GPT-5.6 Terra & high & 1.05M & 120K \\
GPT-5.6 Luna & high & 1.05M & 120K \\
Claude Fable 5 & high & 1.00M & 120K \\
Claude Opus 5 & high & 1.00M & 120K \\
Claude Sonnet 5 & high & 1.00M & 120K \\
GPT-5.6 Sol & xhigh & 1.05M & 120K \\
Claude Opus 5 & xhigh & 1.00M & 120K \\
\bottomrule
\end{tabular}
\end{table}

\paragraph{Resource bounds.}
Coding and gameplay sessions each have a 24-hour wall-clock limit. A generated
workspace or selected snapshot may occupy at most 1 GiB.
Model calls use a 3,600-second retry budget and the output limits in
Table~\ref{tab:conditions}. Coding output caps are 128K for GPT-5.6 and
Claude profiles and 64K for GPT-6 Astra. Exact storage limits, model profiles,
and recovery controls are included in the artifact.
Gameplay inference has no common total-dollar or
model-call cap: expenditure is measured, and the build prompt asks systems to
use inference economically, while the stated wall-clock, action, output, and
storage bounds remain enforced.

\paragraph{\rhae{} definition.}
For a game with levels $\ell=1,\ldots,L$, hidden human action baselines
$b_\ell$, and system actions $a_\ell$, a completed level within the
$5b_\ell$ action budget receives
$s_\ell=\min\{100(b_\ell/a_\ell)^2,115\}$ and other levels receive zero.
With level weights $w_\ell=\ell$, the game score is
\[
\min\!\left\{
\frac{\sum_{\ell=1}^{L}w_\ell s_\ell}{\sum_{\ell=1}^{L}w_\ell},
100\frac{\sum_{\ell=1}^{C}w_\ell}{\sum_{\ell=1}^{L}w_\ell}
\right\},
\]
where $C$ is the completed prefix. Reported split scores average this quantity
over games. The second term caps credit by weighted completion.

\section{Minimal Starting Prompt}
\label{app:prompt}

The build model receives the following instruction. The controller substitutes
only the runtime model's display name and configured reasoning effort. In
particular, the prompt does not prescribe a solver architecture, visual
representation, memory design, planning method, or rule for any game.

\begin{quote}
\small
\textbf{Build a system for solving ARC-AGI-3 games. Work autonomously to
implement, test, and document the complete solution without asking for human
input.}

The system will be evaluated on the tradeoff between RHAE and inference cost.
Aim primarily for high RHAE while using inference economically. Use additional
inference only when it is reasonably expected to produce a meaningful
improvement in RHAE. Inference cost is measured in US dollars according to
\texttt{docs/pricing.md}. RHAE rewards action efficiency. When outcomes have
equal RHAE, fewer game actions are preferred.

At evaluation, the system receives exactly one current game and must infer its
mechanics from observations available during that run. It runs with Python
3.12 without internet access or access to other games. It may use the
configured runtime model only through the supplied
\texttt{llm\_client.py}. Exact model capabilities and limits are in
\texttt{docs/model.md}.

The local rules and interface are in \texttt{docs/arc\_rules.md}. The supplied
files are read-only. Expose a non-interactive
\texttt{./run.sh --game <game-id> --out-dir <directory>} entrypoint.
\end{quote}

\section{Related-System Comparison}
\label{app:related}

Table~\ref{tab:closest-prior} compares the evaluation unit, selection rule, and
editable artifact. Our protocol evaluates each individual descendant of a
frozen parent on held-out tasks that did not influence the update, with no
performance-based selection among qualified descendants.
HUIB separately evaluates models' ability to produce useful updates and to
use updated artifacts in an online task stream \citep{lin2026huib}; here,
one model profile fills both roles.

\begin{table}[H]
\caption{Closest protocol precedents. Strict test means that test tasks do not
influence proposals, acceptance, stopping, or candidate selection.}
\label{tab:closest-prior}
\centering
\scriptsize
\setlength{\tabcolsep}{3.5pt}
\renewcommand{\arraystretch}{1.12}
\begin{tabularx}{\linewidth}{p{0.14\linewidth}X X X}
\toprule
Work & Editable artifact and schedule & Final evidence & Main protocol difference \\
\midrule
STOP & One optimizer function over 1--4 recursive depths & Fresh optimization tasks & Restricted to an optimizer-function artifact. \\
ADAS & Code-defined agent functions under archive-guided search & All discovered ARC agents on held-out test; five evaluations per agent & Searches agent functions; our study tracks prescribed revisions of complete interactive harnesses. \\
AFlow / GEPA & Workflows or prompts under budgeted search & Validation-selected candidate on test & Constrained artifact and selection over many candidates. \\
SICA / DGM & Complete coding-agent codebase with archive-guided evolution & Same-benchmark selection plus transfer & Whole-code evolution with variable lineages and update counts. \\
Meta-Harness / AHE & Executable harness source under multi-iteration score-and-trace search & Best or final frozen harness & Seeded candidate search with selected-harness outcomes. \\
RRSI \citep{xia2026rrsi} & Harness source; 20--40 rounds with regularized proposals and acceptance & Evolved harness on held-out tasks and OOD benchmarks & Tests a selection intervention under multi-candidate search, rather than every prescribed descendant. \\
HarnessCompass & Minimal Bash-only coding harness over five rounds & Final harness on 450 untouched tasks & Close minimal-seed strict test in a different domain. \\
HUIB / SEAGym / Rethinking & Skills, prompts, memory, or seeded harnesses & In-stream or strict-test evaluation & Studies artifact use, regressions, and allocation of rollout budgets. \\
HarnessOpt-Bench & Executable seed under trusted execution & Inaccessible strict test after one nomination & Candidate search and final nomination introduce selection. \\
Workspace Optimization & Structured ARC workspace with online repair within games & Public 25-game suite & Artifact repair occurs within individual games. \\
\textbf{Ours} & \textbf{Complete ARC harness with exactly two revisions} & \textbf{Every snapshot on 5 development and 20 strict held-out games} & \textbf{Prescribed lineages; primary endpoint is the final revision. Joint RHAE, actions, and dollar cost.} \\
\bottomrule
\end{tabularx}
\end{table}

\section{Additional Quantitative Results}
\label{app:quantitative}

\begin{table}[H]
\caption{All completed trajectories reported in the paper. Conditions are
grouped by provider, with matched higher-effort profiles first. Logical costs
use the publication price schedule and selected per-game token ledgers; three
trajectories are recovered directly from those ledgers because
final aggregate report assembly failed.}
\label{tab:replicate-score-cost}
\centering
\setlength{\tabcolsep}{2.0pt}
\renewcommand{\arraystretch}{1.01}
\begin{minipage}[t]{0.485\linewidth}
\vspace{0pt}
\centering
\scriptsize
\begin{tabular}{@{}lrrrr@{}}
\multicolumn{5}{c}{\textbf{OpenAI}} \\
\toprule
& & \multicolumn{3}{c}{\rhae{} / cost (USD/game)} \\
\cmidrule(lr){3-5}
Condition & Rep. & Initial & Revision 1 & Revision 2 \\
\midrule
\multirow{3}{*}{\shortstack[l]{GPT-6 Astra\\high}}
 & 1 & 97.14/11.90 & 93.77/10.86 & 97.55/10.24 \\
 & 2 & 58.03/8.58  & 80.76/12.53 & 92.62/11.96 \\
 & 3 & 93.59/18.77 & 93.23/11.87 & 92.94/12.92 \\
\addlinespace[3pt]
\multirow{3}{*}{\shortstack[l]{GPT-5.6 Sol\\xhigh}}
 & 1 & 43.95/31.39 & 18.49/4.89 & 21.99/7.18 \\
 & 2 & 37.64/12.62 & 56.37/13.09 & 49.86/17.69 \\
 & 3 & 44.36/37.89 & 42.43/23.73 & 53.96/22.36 \\
\addlinespace[2pt]
\multirow{3}{*}{\shortstack[l]{GPT-5.6 Sol\\high}}
 & 1 & 20.38/12.87 & 29.24/8.19 & 53.56/8.15 \\
 & 2 & 27.92/8.84  &  6.82/2.90 & 12.74/5.61 \\
 & 3 & 30.15/25.39 & 26.17/6.26 & 19.45/5.89 \\
\addlinespace[2pt]
\multirow{3}{*}{\shortstack[l]{GPT-5.6 Terra\\high}}
 & 1 & 4.77/4.68 & 0.94/4.70 & 0.49/4.24 \\
 & 2 & 4.39/9.88 & 1.72/1.55 & 1.04/1.70 \\
 & 3 & 3.33/3.89 & 0.60/2.49 & 0.55/3.34 \\
\addlinespace[2pt]
\multirow{3}{*}{\shortstack[l]{GPT-5.6 Luna\\high}}
 & 1 & 1.52/0.46 & 1.71/0.57 & 1.90/0.40 \\
 & 2 & 1.75/0.14 & 1.87/0.44 & 0.75/0.17 \\
 & 3 & 1.09/2.00 & 0.33/0.85 & 0.26/0.77 \\
\bottomrule
\end{tabular}
\end{minipage}
\hfill
\begin{minipage}[t]{0.505\linewidth}
\vspace{0pt}
\centering
\scriptsize
\begin{tabular}{@{}lrrrr@{}}
\multicolumn{5}{c}{\textbf{Anthropic}} \\
\toprule
& & \multicolumn{3}{c}{\rhae{} / cost (USD/game)} \\
\cmidrule(lr){3-5}
Condition & Rep. & Initial & Revision 1 & Revision 2 \\
\midrule
\multirow{3}{*}{\shortstack[l]{Claude Opus 5\\xhigh}}
 & 1 & 27.85/9.18  & 28.79/8.18 & 36.39/12.48 \\
 & 2 & 43.66/13.62 & 46.01/12.76 & 46.82/11.44 \\
 & 3 & 22.29/10.77 & 36.84/8.65 & 36.63/7.75  \\
\addlinespace[2pt]
\multirow{3}{*}{\shortstack[l]{Claude Opus 5\\high}}
 & 1 &  6.57/1.12 &  3.79/2.71 &  2.52/2.46 \\
 & 2 & 16.46/3.93 & 21.47/7.16 & 26.48/9.45 \\
 & 3 & 12.09/6.78 & 20.28/4.30 & 17.28/4.56 \\
\addlinespace[2pt]
\multirow{3}{*}{\shortstack[l]{Claude Fable 5\\high}}
 & 1 & 16.05/10.63 & 18.37/11.26 & 13.77/10.12 \\
 & 2 & 28.35/17.29 & 41.41/15.99 & 31.55/16.98 \\
 & 3 & 18.92/13.81 & 26.26/14.15 & 26.04/12.33 \\
\addlinespace[2pt]
\multirow{3}{*}{\shortstack[l]{Claude Sonnet 5\\high}}
 & 1 & 0.00/0.27 & 0.00/0.33 & 0.11/1.24 \\
 & 2 & 0.12/0.88 & 0.21/1.48 & 0.81/1.70 \\
 & 3 & 0.14/0.49 & 0.52/1.24 & 0.14/1.36 \\
\bottomrule
\end{tabular}
\end{minipage}
\end{table}

Table~\ref{tab:replicate-score-cost} gives the complete held-out score--cost
trajectory for each completed construction, with full model versions and
reasoning-effort profiles.

\paragraph{Construction and qualification outcomes.}
All 27 planned runs in the retained study cohort produce a smoke-qualified
initial system and complete all three evaluations. Paired revision analyses
use all 27 complete lineages. Across their 81 scored stages,
46 qualify the raw candidate without repair and 35 select a smoke-repaired
candidate. Initial held-out \rhae{} averages 24.54 and ranges from 0 to 97.14.

\paragraph{Uncertainty estimation.}
Percentile confidence intervals use 100,000 whole-lineage bootstrap resamples
with seed 20260902 and the artifact's fixed pseudonymous trajectory order.
Each resampled lineage retains all stages and games, preserving paired
comparisons across versions. Intervals condition on the observed study pool
and fixed game split, not new tasks or feedback partitions.

\paragraph{Cost accounting.}
Logical solving cost sums priced usage for fresh and replayed responses in
each suite-selected game attempt; the per-game meters remain auditable
independently of final experiment-report assembly. For cross-model comparison,
dollar values are recomputed from the immutable token ledgers using first-party
standard API list prices frozen on September 13, 2026. Run-time pricing
profiles remain in the artifact.
These dollar figures cover model inference; they exclude infrastructure and
human labor and do not measure latency or energy use.

\begin{table}[H]
\caption{Descriptive held-out means for matched reasoning-effort conditions.
Each row contains three independent trajectories.}
\label{tab:effort}
\centering
\small
\begin{tabular}{lrrr}
\toprule
Condition & Initial & Revision 2 & Change \\
\midrule
Sol high & 26.15 & 28.58 & +2.43 \\
Sol xhigh & 41.98 & 41.94 & -0.05 \\
Opus high & 11.71 & 15.43 & +3.72 \\
Opus xhigh & 31.27 & 39.95 & +8.68 \\
\bottomrule
\end{tabular}
\end{table}

\paragraph{Development-based checkpoint selection.}
As a post hoc diagnostic, selecting the preserved snapshot with highest
five-game development \rhae{} chooses $H_0/H_1/H_2$ in 9/6/12 lineages and
yields 29.06 mean held-out \rhae{}: $+4.53$ over $H_0$ (95\% interval
$[0.81,8.68]$) and $+1.72$ over always deploying $H_2$ ($[0.38,3.67]$).
Mean deployment cost rises from \$7.57 to \$9.15 per game.
The rule uses only development scores and could be automated; its post hoc
evaluation does not establish the performance of a prospectively specified
selection policy (Table~\ref{tab:checkpoint-selection}).

\begin{table}[H]
\caption{Post hoc checkpoint diagnostic. Development selection chooses the
snapshot with highest five-game development \rhae{}, breaking ties toward the
earlier snapshot. ``$+/-/=$'' counts held-out improvement, regression, and no
change relative to $H_0$. Costs exclude outer-loop construction and selection
expenditure. The held-out oracle is diagnostic only.}
\label{tab:checkpoint-selection}
\centering
\scriptsize
\setlength{\tabcolsep}{3.5pt}
\begin{tabular}{lrrrrc}
\toprule
Deployment policy & \rhae{} & Gain & USD/game & $+/-/=$ &
$H_0/H_1/H_2$ \\
\midrule
Retain $H_0$ & 24.54 & 0.00 & 10.30 & 0/0/27 & 27/0/0 \\
Deploy $H_2$ & 27.34 & +2.80 & 7.57 & 15/12/0 & 0/0/27 \\
Select by development \rhae{} & 29.06 & +4.53 & 9.15 & 13/5/9 & 9/6/12 \\
Held-out oracle & 30.66 & +6.13 & 9.80 & 18/0/9 & 9/8/10 \\
\bottomrule
\end{tabular}
\end{table}

\begin{table}[H]
\caption{Logical solving inference cost per held-out game across all 27 scored
lineages. Logical cost includes replayed responses.}
\label{tab:cost-stage}
\centering
\small
\begin{tabular}{lrr}
\toprule
Snapshot & Mean USD/game & Median USD/game \\
\midrule
Initial & 10.30 & 9.18 \\
Revision 1 & 7.15 & 6.26 \\
Revision 2 & 7.57 & 7.18 \\
\bottomrule
\end{tabular}
\end{table}

\begin{table}[H]
\caption{Joint held-out score--cost movements. The ``cheaper/\rhae{} down''
category records joint decreases in cost and \rhae{}.}
\label{tab:cost-movements}
\centering
\scriptsize
\setlength{\tabcolsep}{4pt}
\begin{tabular}{lrrrrrr}
\toprule
Transition & Pareto gain & Score$\uparrow$/cost$\uparrow$ &
Cheaper/\rhae{}$\downarrow$ & Worse both & Tie/cost$\downarrow$ &
Tie/cost$\uparrow$ \\
\midrule
$H_0\!\to\!H_1$ & 6 & 9 & 9 & 2 & 0 & 1 \\
$H_1\!\to\!H_2$ & 6 & 6 & 8 & 7 & 0 & 0 \\
$H_0\!\to\!H_2$ & 9 & 6 & 9 & 3 & 0 & 0 \\
\bottomrule
\end{tabular}
\end{table}

\begin{table}[H]
\caption{Inference cost in USD under the frozen publication price schedule.
Selected-game cost counts fresh model inference; logical replay charges used
for score--cost comparison are excluded. Coding-bootstrap cost covers
construction, revision, and repair; qualification is runtime inference used
to select a smoke-qualified candidate.}
\label{tab:outer-cost}
\centering
\small
\begin{tabular}{lrrrr}
\toprule
Component & Initial & Revision 1 & Revision 2 & Total \\
\midrule
Coding loop, 27 scored & 607.92 & 371.78 & 372.64 & 1,352.34 \\
Smoke qualification, 27 planned & 408.02 & 111.11 & 127.72 & 646.84 \\
\midrule
Fresh selected-game attempts, 27 scored & 6,749.95 & 4,712.60 & 4,667.55 & 16,130.09 \\
\textbf{List-price-equivalent total} & \textbf{7,765.89} & \textbf{5,195.49} & \textbf{5,167.90} & \textbf{18,129.27} \\
\bottomrule
\end{tabular}
\end{table}

The 27 lineages ran over 29 days (17 August to 15 September 2026) across two
machines; with pilot and replaced runs, the experimental program spanned at
least five weeks. It corresponds to at least \$18,129.27 under the frozen
publication price schedule. Fresh selected-game attempts account for \$16,130.09. Across the 27
scored lineages, coding-bootstrap inference for construction, revision, and
repair totals at least \$1,352.34.
Smoke qualification adds \$646.84 across all planned runs.

We report endpoint score and solving cost jointly: nine descendants are
strict deployment score--cost Pareto gains, while nine apparent savings are
performance contractions.
Where aggregate report accounting is unavailable, logical cost is recovered
from immutable suite selections and complete per-game meters. Unselected
game-attempt recovery overhead is outside the reported program lower bound
(Appendix~\ref{app:integrity}). Coding cost contains three unresolved requests, all in revision~1;
qualification contains seven. We do not impute these requests, so
Table~\ref{tab:outer-cost} reports lower bounds. The same treatment applies to
one further unresolved provider request outside these two cost components.

\paragraph{Human-steered reference systems.}
\label{app:human-designed}
Human steering refers to development decisions and is compatible with
AI-written code.
Tycho reports 100.00 \rhae{} over all 25 public games with GPT-5.6 Sol
at maximum reasoning effort and with Opus 5 at xhigh
\citep{lehmann2026tycho}. VISTA and AVO also report 100.00 with Opus 5
\citep{han2026vista,chen2026avoarc}; VISTA uses xhigh and reports 98.27
for its separate Sol-max run. Retrodict reports 99.86 with Sol at maximum
effort \citep{brown2026retrodict}. In our study, revision-2 held-out means are
28.58/41.94 for Sol at high/xhigh and 15.43/39.95 for Opus at high/xhigh
(Table~\ref{tab:effort}).
The Opus comparison matches both model version and reasoning effort.
Tycho designates the same five development games as our study
(\texttt{tr87}, \texttt{vc33}, \texttt{r11l}, \texttt{bp35}, \texttt{ft09})
and scores 100 on each of the twenty remaining games. The performance gap
therefore persists on the same evaluation games. The Sol comparison uses
different reported effort labels (max versus high/xhigh).

The development and selection procedures differ.
Tycho selects among four policies using Opus 4.8 on all 25 public games,
then evaluates the selected policy with Opus 5 and Sol; our twenty held-out
games provide no feedback for construction, revision, or selection.
Retrodict likewise reports earlier development attempts on all public games.
Reported costs vary substantially across these systems and use different
pricing dates and accounting conventions; we do not normalize their
historical costs to our frozen publication price schedule.
Inference allocation differs as well. Tycho reports about 15,100 model calls
with Opus 5 and 26,500 with Sol over the 25 games, roughly 600 and 1,060 per
game. Our revision-2 harnesses average 13.9 (Opus high), 18.1 (Opus xhigh),
78.7 (Sol high), and 114.2 (Sol xhigh) logical model responses per held-out
game. The two counts are instrumented differently, so the contrast is one of
order of magnitude.
Call counts are not the only difference, and spend alone does not account
for the gap: Retrodict reports about \$26 per game at its list prices, less
than two of our initial Sol-xhigh harnesses spent (\$31--38 per game for
about 44 \rhae{}; Table~\ref{tab:replicate-score-cost}).
The comparison establishes attainable performance with these models, but
does not isolate the contribution of human steering or establish equivalent
efficiency.

The coding-agent baselines cited in the introduction come from within-study
comparisons. Schema reports 42.83 \rhae{} for its generic Claude Code baseline
and 98.98 for its world-model harness using the same Opus 4.8/Fable 5 pairing
\citep{zeng2026schema}. Twin reports 61.1 for off-the-shelf Codex with Sol and
93.3 with its world-model harness, holding the model, file bridge, and sandbox
fixed; the stronger harness also uses more inference compute
\citep{skoutnev2026twin}.

\begin{table}[H]
\caption{Endpoint robustness to practical \rhae{} thresholds. Changes within
$\pm\epsilon$ are indeterminate; deployment Pareto gains also require lower
solving cost.}
\label{tab:threshold-sensitivity}
\centering
\small
\begin{tabular}{rrrrr}
\toprule
$\epsilon$ & Improve & Regress & Indeterminate & Pareto gain \\
\midrule
0 & 15 & 12 & 0 & 9 \\
1 & 11 & 8 & 8 & 7 \\
5 & 9 & 3 & 15 & 5 \\
\bottomrule
\end{tabular}
\end{table}

\subsection{Fixed-harness execution variance}
\label{app:fixed-harness-variance}

The study evaluates each frozen snapshot once. To measure
run-to-run execution variance separately from construction and revision, we
reran one snapshot after the campaign: Sol-high replicate~1 at $H_2$. The
snapshot was selected post hoc after campaign outcomes were available, and
GPT-5.6 was chosen for its operational speed; this is a diagnostic case rather
than a random sample of harnesses. The original evaluation ran in August 2026
and the nine fresh evaluations in September 2026. Recorded harness,
controller, model-profile, provider-configuration, and environment fingerprints
are identical across evaluations, although unexposed provider-side changes
cannot be excluded. Each evaluation uses the same twenty held-out games and
starts new runtime-model inference. No construction, revision, repair,
checkpoint selection, or hosted replay occurs in the added runs.

\begin{table}[H]
\caption{Ten evaluations of one unchanged Sol-high $H_2$ harness. Evaluation
0 is the original campaign observation. ``Complete'' counts games for which
all levels were solved; \rhae{} also gives partial credit within the remaining
games.}
\label{tab:fixed-harness-variance}
\centering
\small
\begin{tabular}{rrrrr}
\toprule
Evaluation & \rhae{} & Actions & Levels completed & Complete games \\
\midrule
0 & 53.56 & 8,492 & 102 & 8 \\
1 & 46.86 & 8,509 & 94 & 6 \\
2 & 43.38 & 9,382 & 90 & 6 \\
3 & 48.21 & 9,389 & 92 & 7 \\
4 & 45.32 & 8,515 & 91 & 6 \\
5 & 55.37 & 8,459 & 104 & 8 \\
6 & 46.52 & 9,435 & 99 & 7 \\
7 & 49.23 & 9,410 & 92 & 6 \\
8 & 52.10 & 8,648 & 96 & 8 \\
9 & 44.75 & 10,135 & 93 & 5 \\
\bottomrule
\end{tabular}
\end{table}

Across all ten evaluations, mean \rhae{} is 48.53, the sample standard
deviation is 3.99, and the range is 43.38--55.37. These are descriptive
statistics conditional on the post hoc snapshot. The nine fresh evaluations
alone have mean 47.97 and sample standard deviation 3.79, so the observed
variability does not depend on including the older campaign observation. The
original 53.56 observation is the second highest and lies 5.03 points above
the ten-evaluation mean. The mean absolute difference between two evaluations
is 4.73 points.
Every evaluation remains above this trajectory's 20.38 initial score, so its
endpoint improvement classification is unchanged. Replacing only its original
$H_2$ observation by the ten-run mean changes the campaign-wide endpoint mean
from $+2.80$ to $+2.62$; this is a mechanical sensitivity check rather than a
measurement-error correction for the other trajectories.

Variation is larger at game level: the median within-game sample standard
deviation is 18.78 \rhae{} points, and the three largest are 30.76 on
\texttt{sb26}, 28.33 on \texttt{m0r0}, and 27.13 on \texttt{tn36}. Averaging
twenty games therefore substantially stabilizes the suite score, while
single-trajectory differences of a few points remain difficult to interpret
precisely. The observed 3.99-point suite standard deviation gives empirical
scale to the $\epsilon=5$ sensitivity in
Table~\ref{tab:threshold-sensitivity}; one harness is insufficient to define a
universal noise threshold. These post hoc evaluations are separate from the
construction-and-revision study, and their spend is excluded from
Table~\ref{tab:outer-cost}.
The supplement releases all 200 evaluation--game rows and a standalone
reproduction command for these statistics.

\paragraph{Weighting and game sensitivity.}
The endpoint mean is $+2.80$ under both trajectory and condition weighting,
since each condition has three complete lineages, and $+2.55$ when the seven
model families receive equal weight. 
The condition-stratified bootstrap interval is $[-1.12,7.01]$ under either weighting and remains inconclusive.
Leaving out one condition at a time yields means from $+1.72$ to $+3.59$ and adaptation gaps from 4.07 to 6.40.
Across held-out games, 13 have a positive mean endpoint change and seven a negative one; the largest gain is $+20.44$ on \texttt{lp85}, while the largest decline is $-15.03$ on \texttt{sb26}. 
The small pooled gain is distributed across conditions and concentrated in opposing game-level effects. 
These intervals condition on the fixed games and do not quantify uncertainty over new development partitions.

Omitting \texttt{r11l} only from the observed development summary reduces the
mean development--held-out gap from 5.62 to 3.02 points. The corresponding
gaps after omitting each of the other development games range from 4.68 to
7.84. This is a descriptive influence check, not a counterfactual experiment:
all harnesses were still revised using the original five-game feedback.
Tables~\ref{tab:completion-context} and~\ref{tab:game-influence} give completion,
action, and per-game context. Higher level completion accompanies the pooled
score gain, but aggregate action counts alone do not distinguish improved
efficiency from earlier stopping. The supplement reproduces these summaries
from the same published game rows.

\begin{table}[H]
\caption{Held-out completion and action context across the same 540 evaluations
per stage (27 lineages $\times$ 20 games). Level completion is the mean
within-game fraction of completed levels. Actions are mean charged actions;
lower action counts need not imply better efficiency because systems may stop
without completing a game. These summaries do not isolate the components of \rhae{}.}
\label{tab:completion-context}\centering\small
\begin{tabular}{lrrrr}\toprule
Stage & \rhae{} & Complete games & Levels completed & Actions \\\midrule
Initial & 24.54 & 73/540 (13.5\%) & 34.5\% & 259.6 \\
Revision 1 & 25.87 & 83/540 (15.4\%) & 36.4\% & 292.0 \\
Revision 2 & 27.34 & 94/540 (17.4\%) & 37.1\% & 273.6 \\
\bottomrule\end{tabular}\end{table}

\begin{table}[H]
\caption{Per-game initial-to-revision-2 changes, averaging the same 27 paired
lineages. Completion columns are percentage-point changes in complete games
and mean fraction of levels completed. Rows describe observed outcomes;
no alternative feedback split was used to generate the harnesses.}
\label{tab:game-influence}\centering\small
\begin{tabular}{llrrrr}\toprule
Game & Split & $\Delta$\rhae{} & $\Delta$Complete & $\Delta$Levels & $\Delta$Actions \\\midrule
\texttt{bp35} & dev & +3.43 & +3.7 & +5.3 & +46.5 \\
\texttt{ft09} & dev & -0.45 & +11.1 & -2.5 & -2.1 \\
\texttt{r11l} & dev & +18.83 & +14.8 & +17.9 & +22.1 \\
\texttt{tr87} & dev & +8.11 & +7.4 & +11.1 & +22.8 \\
\texttt{vc33} & dev & +12.19 & +11.1 & +13.2 & +76.1 \\
\texttt{ar25} & heldout & -1.29 & +3.7 & -4.2 & -80.7 \\
\texttt{cd82} & heldout & +3.11 & +3.7 & +1.9 & -1.0 \\
\texttt{cn04} & heldout & +1.74 & +3.7 & +0.0 & -39.9 \\
\texttt{dc22} & heldout & -2.39 & +3.7 & -4.3 & +21.2 \\
\texttt{g50t} & heldout & -1.42 & +0.0 & -0.5 & -4.5 \\
\texttt{ka59} & heldout & -1.46 & +7.4 & -1.6 & -13.0 \\
\texttt{lf52} & heldout & -3.75 & -3.7 & -4.4 & +41.1 \\
\texttt{lp85} & heldout & +20.44 & +29.6 & +13.0 & -24.1 \\
\texttt{ls20} & heldout & +4.96 & +3.7 & +7.4 & +87.0 \\
\texttt{m0r0} & heldout & -1.49 & -3.7 & +0.0 & -49.6 \\
\texttt{re86} & heldout & +1.38 & +3.7 & -0.5 & -3.0 \\
\texttt{s5i5} & heldout & +8.25 & +7.4 & +8.3 & -33.6 \\
\texttt{sb26} & heldout & -15.03 & -18.5 & -13.0 & +5.0 \\
\texttt{sc25} & heldout & +1.64 & +0.0 & +2.5 & +46.6 \\
\texttt{sk48} & heldout & +4.69 & +3.7 & +5.1 & +72.4 \\
\texttt{sp80} & heldout & +10.01 & +11.1 & +11.7 & +74.6 \\
\texttt{su15} & heldout & +4.82 & +0.0 & +7.4 & +37.8 \\
\texttt{tn36} & heldout & +4.49 & +7.4 & +5.3 & +44.1 \\
\texttt{tu93} & heldout & +15.66 & +14.8 & +16.9 & +59.7 \\
\texttt{wa30} & heldout & +1.71 & +0.0 & +1.2 & +39.9 \\
\bottomrule\end{tabular}\end{table}

\section{Harness and Strategy Audit}
\label{app:case-studies}

We inspect every selected snapshot, exact diff, updater record, and relevant
smoke-repair artifact in all 27 trajectories.
The strategy labels are
post hoc descriptions derived with outcome visibility. \emph{Preserve} (P)
means repair or tuning without changing the operating strategy. \emph{Extend}
(E) adds a material capability within the same decision loop.
\emph{Reorient} (R) changes control priority, context continuity, replay, or
action/inference scheduling while retaining the broader harness family.
\emph{Replace} would change that family; no transition does so.

\begin{figure}[t]
\centering
\includegraphics[width=\linewidth]{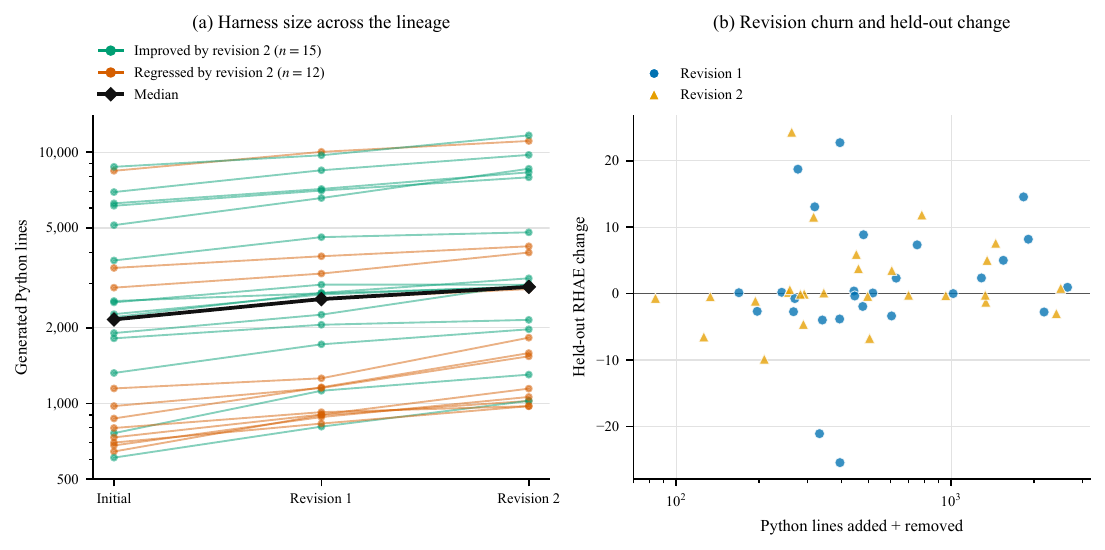}
\caption{\textbf{Harnesses grow and undergo substantial revision without
reliable gains.} Generated Python counts model-produced source and tests and
excludes the supplied \texttt{llm\_client.py}. (a) All 27 lineages on a
log scale, colored by initial-to-revision-2 held-out outcome; the black path is
the stage median. (b) Each point is one selected parent-to-descendant Python
diff. Churn is added plus removed lines, and points above zero improve held-out
\rhae{}.}
\label{fig:harness-evolution}
\end{figure}

Every lineage grows from the initial to the second-revision snapshot.
Median size rises from 2,161 to 2,604 to 2,912 generated Python lines, and the
mean endpoint increase is 1,052 lines. Both large and small diffs lie on each
side of zero held-out change. Edit volume establishes substantive revision;
held-out change determines whether that revision improved performance.

\begin{table}[H]
\caption{Strategy-transition counts across all conditions. Entries give
Preserve/Extend/Reorient.}
\label{tab:strategy-counts}
\centering
\scriptsize
\setlength{\tabcolsep}{3.5pt}
\renewcommand{\arraystretch}{1.12}
\begin{tabularx}{\linewidth}{p{0.13\linewidth}Xrr}
\toprule
Condition & Initial harness family & $H_0\!\to\!H_1$ P/E/R &
$H_1\!\to\!H_2$ P/E/R \\
\midrule
Astra high & Persistent Python workers with policy reuse & 0/0/3 & 0/1/2 \\
Sol high & Direct multimodal batched planner & 0/0/3 & 0/3/0 \\
Terra high & Compact visual/text planner & 0/0/3 & 0/2/1 \\
Luna high & Direct model controller & 0/2/1 & 0/2/1 \\
Fable high & Strategist, executor, and replay system & 1/0/2 & 1/0/2 \\
Opus high & Layered world-model/planner hybrids & 0/0/3 & 0/0/3 \\
Sonnet high & Heuristic/model hybrid & 0/1/2 & 1/2/0 \\
Sol xhigh & Multimodal batched planner & 0/1/2 & 1/2/0 \\
Opus xhigh & World-model and replay hybrids & 0/0/3 & 0/0/3 \\
\midrule
All conditions & & 1/4/22 & 3/12/12 \\
\bottomrule
\end{tabularx}
\end{table}

\begin{table}[!htbp]
\caption{Strategy transitions for all 27 replicates. P, E, and R follow the
definitions above; Table~\ref{tab:replicate-score-cost} gives the corresponding
score--cost trajectories.}
\label{tab:strategy-ledger}
\centering
\scriptsize
\setlength{\tabcolsep}{4.5pt}
\renewcommand{\arraystretch}{1.05}
\begin{tabular}{lrcc}
\toprule
Condition & Replicate & $H_0\!\to\!H_1$ & $H_1\!\to\!H_2$ \\
\midrule
Astra high & 1 & R & R \\
Astra high & 2 & R & R \\
Astra high & 3 & R & E \\
Sol high & 1 & R & E \\
Sol high & 2 & R & E \\
Sol high & 3 & R & E \\
Terra high & 1 & R & E \\
Terra high & 2 & R & E \\
Terra high & 3 & R & R \\
Luna high & 1 & E & E \\
Luna high & 2 & E & R \\
Luna high & 3 & R & E \\
Fable high & 1 & R & P \\
Fable high & 2 & P & R \\
Fable high & 3 & R & R \\
Opus high & 1 & R & R \\
Opus high & 2 & R & R \\
Opus high & 3 & R & R \\
Sonnet high & 1 & R & E \\
Sonnet high & 2 & E & P \\
Sonnet high & 3 & R & E \\
Sol xhigh & 1 & R & E \\
Sol xhigh & 2 & E & P \\
Sol xhigh & 3 & R & E \\
Opus xhigh & 1 & R & R \\
Opus xhigh & 2 & R & R \\
Opus xhigh & 3 & R & R \\
\bottomrule
\end{tabular}
\end{table}

\paragraph{First revisions reorganize; second revisions diversify.}
First revisions most often change the operating strategy itself: models alter
context persistence, replay flow, model-consultation cadence, action batching,
or budget allocation. Second revisions split equally between extending
capabilities and reorienting the strategy. Even changes described as
incremental by their builders can change when the runtime model is consulted
or how scarce actions and dollars are scheduled.

\paragraph{Replicates disagree within the same condition.}
The strategy paths are identical across all Sol-high replicates but their
held-out outcomes range from a 33.18-point gain to a 15.18-point loss. Fable,
Luna, Sonnet, and Sol-xhigh also mix preserve, extend, and reorient transitions
within one condition. Opus-high and Opus-xhigh reorient at every stage, yet
their outcomes include both gains and regressions. Strategy change is
therefore common, and its presence or coarse type does not certify improvement.

\paragraph{An evidence-linked improvement.}
Sol-high replicate~1 (T001), selected post hoc for illustration, connects
feedback to a substantial gain. Three development games exhausted $H_1$'s
self-imposed 100-decision limit; its meters recorded repeated cache writes
with few reads. Revision~2 raised this internal limit to 120, removed explicit
cache writes, and added animation-aware observations, object geometry, and
completed-level memory. The pre-edit plan names these changes; the transcript
records 17 passing local tests, and the candidate qualified without repair.
Held-out \rhae{} rose from 29.24 to 53.56 while inference cost per game changed
from \$8.19 to \$8.15. Several components changed together, so the case
illustrates evidence-grounded adaptation with observed transfer without
isolating a causal mechanism. The supplement's \texttt{case-study/} directory
links feedback excerpts, the plan, source, tests, and outcomes.

\paragraph{Cost reduction with performance loss.}
In Sol-xhigh replicate~1 (T007), the first revision replaced response-ID
chaining and history compaction with independent calls carrying a bounded
notebook and recent outcome. Across the two revisions, held-out \rhae{}
fell from 43.95 to 21.99 and inference cost from \$31.39 to \$7.18 per game;
development \rhae{} also fell, from 37.11 to 31.73. Revision~2 partially
recovered both scores. Several components changed together, so any
attribution of the regression to lost contextual information remains
untested. The supplement's \texttt{case-study/T007.md} links the source
changes and per-stage outcomes.

\section{Capability Codebook and Sampling}
\label{app:codebook}

The process-coding corpus covers all 54 parent-to-descendant revision episodes
from the 27 study lineages. Each packet receives two independent annotations
under the same frozen codebook, with model identity and downstream outcomes
masked.

To relate observable revision behavior to held-out outcomes, we analyze
visible model messages, tool calls and results,
generated source, exact diffs, tests, fixed feedback packets, smoke records,
runtime traces, actions, scores, and costs. We do not treat provider-hidden
chain-of-thought as evidence of cognition.

\paragraph{Evidence boundary and units.}
Direct trace evidence is a logged event or immutable artifact. Operational
relations are reproducible links, such as a feedback item followed by a
targeted edit and relevant test. Claims about latent beliefs or understanding
are interpretations and are excluded from quantitative coding. Complete
trajectories are the independent units; events, games, and stages are nested
observations.

\paragraph{Codebook organization.}
The codebook measures four stages of the update process:
\begin{enumerate}
    \item feedback-evidence access, diagnostic specificity, and discriminating
    probes.
    \item plan enactment, feedback coverage, edit alignment, and change scope.
    \item incumbent protection, updater verification, and proximal result.
    \item post-result adaptation and resource allocation.
\end{enumerate}

\begin{table}[H]
\caption{Episode codebook used for observable harness revision.}
\label{tab:capability-codebook}
\centering
\scriptsize
\setlength{\tabcolsep}{3.5pt}
\renewcommand{\arraystretch}{1.12}
\begin{tabularx}{\linewidth}{p{0.18\linewidth}p{0.25\linewidth}X}
\toprule
Code group & Central question & Observable indicators \\
\midrule
Evidence access & Does the updater inspect item-specific support before editing? & Feedback summaries, traces, logs, score breakdowns, errors, and reproducers (\textsc{Evid}). \\
Diagnosis & Does it connect cited evidence to a mechanism and prediction? & Pre-edit symptoms, localized mechanisms, predictions, counterfactuals, and boundary conditions (\textsc{Diag}). \\
Diagnostic probing & Does evidence discriminate among hypotheses? & Targeted inspection, stated alternatives, refutable expectations, and result-governed narrowing (\textsc{Probe}). \\
Plan and alignment & Are targets, actions, criteria, and retained edits linked? & Enacted plans, eligible-feedback coverage, and final-edit alignment (\textsc{Plan}, \textsc{Cov}, \textsc{Align}). \\
Change control & Does it contain risk and protect the incumbent? & Executable scope, fallbacks, baseline checks, comparison, narrowing, rollback, and no-change decisions (\textsc{Scope}, \textsc{Protect}). \\
Verification & Does it test the retained change and remain runnable? & Reproducers, targeted and regression tests, smoke repair, requalification, and proximal results (\textsc{Verify}, \textsc{Result}). \\
Post-result adaptation & Does a post-edit result govern the next action? & Interpretation followed by handoff, edit, revert, changed test, narrowed scope, or criterion-based stop (\textsc{Adapt}). \\
Resource adaptation & Does evidence govern computation allocation? & Calls, context, batching, caching, tests, stopping, numeric triggers, and expected-value comparisons (\textsc{Meta}). \\
\bottomrule
\end{tabularx}
\end{table}

\paragraph{Codebook refinement and consensus.}
An initial blind calibration exposed temporal ambiguities. We tightened only
the pre-edit and post-result anchors; every packet was then coded under frozen
version~2 before its outcome linkage was opened. Consensus labels were frozen by a
conservative rule: exact agreement stands, ordinal disagreement receives the
lower evidenced capability level, and any missing interval remains missing.

\paragraph{Ordinal episode coding.}
The twelve variables are feedback-evidence access (\textsc{Evid}), diagnostic
specificity (\textsc{Diag}), discriminating probe strength (\textsc{Probe}),
plan enactment (\textsc{Plan}), eligible-feedback coverage (\textsc{Cov}),
final-edit alignment (\textsc{Align}), scope containment (\textsc{Scope}),
incumbent protection (\textsc{Protect}), updater verification
(\textsc{Verify}), proximal operational result (\textsc{Result}), post-edit
result-grounded adaptation (\textsc{Adapt}), and resource adaptation
(\textsc{Meta}). Sequential codes credit only evidence visible before the
relevant action. Final summaries can corroborate but cannot create a prior
diagnosis, prediction, probe, plan, or resource criterion. Mandatory final
smoke passage contributes to \textsc{Result}, but raises \textsc{Verify} only
when a failure causes repair and requalification.

\begin{table}[!htbp]
\caption{Unadjudicated reliability under frozen codebook version~2. $n$
excludes variable-specific missingness. $\kappa_w$ uses quadratic disagreement
weights $d_{ij}=((i-j)/(K-1))^2$ for $K$ ordered categories.
$\alpha$ uses nominal disagreement and exposes prevalence effects. Dashes
denote undefined chance-corrected coefficients for invariant labels.}
\label{tab:capability-reliability}
\centering
\scriptsize
\setlength{\tabcolsep}{4pt}
\begin{tabular}{lrrrr}
\toprule
Code & $n$ & Exact & $\kappa_w$ & $\alpha$ \\
\midrule
\textsc{Evid} & 54 & 1.00 & -- & -- \\
\textsc{Diag} & 30 & .67 & .35 & .38 \\
\textsc{Probe} & 53 & .85 & .66 & .67 \\
\textsc{Plan} & 27 & .85 & .70 & .70 \\
\textsc{Cov} & 54 & 1.00 & -- & -- \\
\textsc{Align} & 54 & 1.00 & -- & -- \\
\textsc{Scope} & 54 & .93 & .96 & .88 \\
\textsc{Protect} & 54 & .74 & .47 & .47 \\
\textsc{Verify} & 54 & 1.00 & -- & -- \\
\textsc{Result} & 54 & 1.00 & -- & -- \\
\textsc{Adapt} & 54 & .93 & .84 & .84 \\
\textsc{Meta} & 35 & .71 & .29 & .17 \\
\bottomrule
\end{tabular}
\end{table}

\begin{table}[!htbp]
\caption{Exploratory process associations for labels with adequate
inter-session reliability and outcome coverage. Contrasts
are immediate held-out \rhae{} change at the highest minus lowest shown code.
Intervals resample complete trajectories; $p$ values permute trajectory profiles.
Associations are descriptive.}
\label{tab:process-results}
\centering
\scriptsize
\setlength{\tabcolsep}{3.5pt}
\begin{tabularx}{\linewidth}{lrrrX}
\toprule
Code & $n$ & $\kappa_w$ & Contrast/counts & Interpretation \\
\midrule
Scope containment & 54 & .96 & $-3.94$ & Scoped minus broad, interval $[-9.09,1.22]$, $p=.12$. \\
Post-result adaptation & 54 & .84 & $-3.24$ & Further-action minus handoff, interval $[-8.54,2.38]$, $p=.174$. \\
\bottomrule
\end{tabularx}
\end{table}

\paragraph{Outcome linkage.}
Coder files and conservative consensus labels were hashed before opening the
private episode key. Associations use immediate parent-to-child held-out change.
Confidence intervals resample trajectories. Permutation tests shuffle complete
two-stage outcome profiles across trajectories while preserving stage.
Our exploratory inclusion rule requires $\kappa_w\geq .67$; codes below this
cutoff and invariant high-agreement codes are reported descriptively.
The artifact provides aggregate reliability,
prevalence, outcome, and association tables and fixes 10,000 resamples and
permutations per process association with deterministic seeds. These tests
are exploratory and unadjusted for multiple comparisons. Shuffling across
conditions assumes exchangeable trajectory profiles, which condition-level
differences may violate; the reported $p$ values do not remove confounding.

\paragraph{Planning-label missingness.}
The planning interval is observable in 27 packets and absent in 27. Missing
packets average 9.22 executable files changed and 1,147 lines added, compared
with 3.74 files and 414 lines when planning is observed. Missingness is also
condition-dependent: all Opus-high, Opus-xhigh, and Sonnet packets are missing,
whereas all Astra, Luna, and Terra packets are observed. It is almost balanced
between revision stages (13 missing at revision~1 and 14 at revision~2).
Consequently, the reliable agreement coefficient for the observed subset does
not justify a planning--outcome conclusion.

\paragraph{Reactive-episode context.}
Episodes with further result-driven action start from higher-scoring parents
on average (32.94 versus 21.33 held-out \rhae{}) and use more updater sessions
(1.67 versus 1.11). Within the ten trajectories whose two revisions receive
different adaptation labels, the reactive revision has lower change in eight.
These descriptive differences do not isolate the effect of repair from the
problems that prompted it.

\section{Campaign Integrity and Recovery Audit}
\label{app:integrity}

The experimental protocol and coding bootstrap remained fixed throughout
the campaign. A post-collection source audit confirms unchanged prompt,
feedback, qualification, game-interface, scoring, and resource-control
implementations across the recorded versions. Operational recovery and
provenance maintenance account for changes in execution-support code; the
artifact supplies a concise source change log. Retained-record checks cover
all 27 complete submitted lineages. Table~\ref{tab:integrity} summarizes these checks;
the supplementary artifact provides pseudonymous audit exports and an offline
consistency checker. These checks establish preservation and consistency of
recorded controls, not an independent experimental rerun.

\begin{table}[H]
\caption{Post-collection integrity checks against retained original records.}
\label{tab:integrity}
\centering
\small
\begin{tabular}{p{0.25\linewidth}p{0.67\linewidth}}
\toprule
Control & Verified result \\
\midrule
Frozen source & All 81 original snapshot fingerprints and file counts match,
covering 1,985 files. \\
Evaluated identity & All 2,025 selected evaluations match the corresponding
snapshot, runtime model-profile fingerprint, environment fingerprint, and
configured output cap. \\
Prompts and bounds & Each rendered prompt is constant across replicates
within each of nine conditions; all 27 lineages share one resource-limit
profile. \\
Hosted validation & All 81 validations are complete, cover 25 games, agree
with recorded scores, match snapshot identity, and use the same trace
identity as local validation. \\
Coding preservation & All 54 blinded packets match their manifests;
their labels and rationales are unchanged from the hash-verified frozen
coding passes and consensus. \\
\bottomrule
\end{tabular}
\end{table}

\paragraph{Execution recovery.}
Construction replacement and operational recovery are distinct. Recovery
finishes an evaluation of the same frozen harness, using compatible journals
for resumption when available. The controller also permits a fresh execution
without a replay parent; such a restart can draw new model responses and is
not deterministic resumption.

The selected-suite manifests record 2,130 attempts for 2,025 game--snapshot
slots across 162 suites. Sixty-three slots have multiple attempts. Each slot
contains exactly one clean attempt, which is selected; the records show no
comparison of multiple clean completed scores to select a higher one. Of the
105 other attempts, 49 are classified as provider infrastructure failures,
11 as blocked, and 45 as incomplete without an outcome reason.
Twenty-one slots contain a fresh restart; seventeen selected attempts are
those fresh restarts. No resource-stop record was found among these 2,130
attempts. These classifications do not fully establish the cause of every
interruption. The released audit contains the attempt histories and detailed
categories. It covers manifest-listed histories and does not estimate a causal
effect of recovery on scores.

\paragraph{Recovery costs.}
Endpoint solving cost includes fresh and replayed inference in the selected
attempt. The program lower bound sums publication-priced fresh selected-game
inference, observed construction/revision/repair inference, and smoke
qualification. It excludes unselected game-attempt recovery overhead;
unresolved terminal requests are not imputed. Operational recovery meters
have not been fully reconciled to the publication pricing schedule. The
reported lower bound should therefore not be interpreted as an exhaustive
bill for the campaign.

\paragraph{Bounded frame exception.}
The validator accepts a narrowly specified frame-hash discrepancy caused by
unseeded visual noise in \texttt{lf52}, while requiring agreement on state,
level completion, and available actions. The exact predicate and expected
and observed frame hashes are retained in the audit. Hosted
validations contain 213 such frame events, all on \texttt{lf52}; local
validation records the same event count. All 81 hosted validations report
score agreement. This exception concerns visual observations, not a relaxed
score criterion.

\paragraph{Auditable process examples.}
The supplement includes the frozen per-code rationales for both coding
passes and consensus across all 54 episodes. Three excerpts were selected
by fixed rules using packet IDs and labels, without downstream outcomes, to
illustrate agreement, disagreement, and missing evidence. Each includes
ordered evidence anchors, labels, rationales, and the consensus rule.
Selection rules and compression markers are preserved in the artifact.
These examples support inspection of the coding method; they are not complete
transcripts or human validation.

\paragraph{Artifact availability.}
\label{app:artifact}
The supplementary
{} artifact contains the seed repository, frozen
prompts and splits, condition, model, and pricing profiles, the scoring and
feedback-curation code, all 81 anonymized source
snapshots with original and review-copy fingerprints, trajectory and game
tables, coder and consensus labels with their rationales, three chronological
coding examples, two revision case studies, integrity and recovery audits,
the capability manual, deterministic analyses, and a checksum manifest.
Offline checks verify configuration, source fingerprints, audit consistency,
labels, and example selection. The release supports reproducing the reported
analyses, inspecting harness changes, and testing new methods for autonomous
improvement.
The artifact is provided as
\texttt{harness-adaptation-review-artifact.zip} in the ancillary files accompanying
this arXiv version.

\section{Additional Analyses}
\label{app:discussion}
This appendix collects  additional analyses. New
bootstrap intervals use seed 20261108 and may differ from the main-text
intervals by a few hundredths.

\subsection{Scale of the development signal}
The five development games are multilevel interactive environments
(Table~\ref{tab:dev-scale}). A failed level ends in \texttt{GAME\_OVER} and can
be reset any number of times, but each reset is charged to that level's
$5b_\ell$ budget; there is no fixed number of lives and no whole-game action
cap. The 54 feedback bundles that seed the revisions summarize 270
parent-game evaluations ($H_0$ for revision~1, $H_1$ for revision~2), with
883 observed and 666 completed level instances and about 42,700 scored
actions, from parent runs that made 12,347 model calls; the bundles,
including traces, frames, and costs, total 352~MiB. Level instances repeat
levels across harnesses, and not every harness reached every level. The 20 held-out games contain 149 levels and
receive 1,620 evaluations, none of which informs construction, revision, or
selection.

\begin{table}[H]
\caption{Development games. Baselines are the hidden human action counts per
level; the controller charges at most $5b_\ell$ actions to level $\ell$. The
last column is the most one game evaluation can charge if every level
exhausts its budget; the total row gives the five-game suite.}
\label{tab:dev-scale}
\centering
\small
\begin{tabular}{lrrrr}
\toprule
Game & Levels & Baseline per level & $5\times$ budget per level &
Max.\ charged per evaluation \\
\midrule
\texttt{tr87} & 6 & 40--146 & 200--730 & 2,070 \\
\texttt{vc33} & 7 & 7--152 & 35--760 & 2,235 \\
\texttt{r11l} & 6 & 22--52 & 110--260 & 1,165 \\
\texttt{bp35} & 9 & 21--163 & 105--815 & 3,255 \\
\texttt{ft09} & 6 & 12--65 & 60--325 & 1,040 \\
\midrule
All five & 34 & 7--163 & 35--815 & 9,765 \\
\bottomrule
\end{tabular}
\end{table}

\subsection{Model versions and run dates}
Table~\ref{tab:model-dates} lists the model version, reasoning effort, and
run dates of each condition. Within a condition, the same model and effort
serve construction, revision, and gameplay, and all three lineages share one
model profile.

\begin{table}[H]
\caption{Model versions and run dates. Dates span the first construction
start to the last recorded controller transition among a condition's three
lineages. All models were accessed through a hosted inference API that
exposed no dated model versions, so the dates give the access window;
provider-side updates within it cannot be excluded. The controller fixes the
model and reasoning effort on every coding and gameplay request; neither the
controller nor any released harness sets a sampling temperature.}
\label{tab:model-dates}
\centering
\small
\begin{tabular}{lllr}
\toprule
Condition & Model version & Effort & Run dates (2026, UTC) \\
\midrule
GPT-6 Astra & \texttt{gpt-6-astra} & high & 13--15 Sep \\
GPT-5.6 Sol & \texttt{gpt-5.6-sol} & high & 17--21 Aug \\
GPT-5.6 Terra & \texttt{gpt-5.6-terra} & high & 17--21 Aug \\
GPT-5.6 Luna & \texttt{gpt-5.6-luna} & high & 17--21 Aug \\
Claude Fable 5 & \texttt{claude-fable-5} & high & 17--22 Aug \\
Claude Opus 5 & \texttt{claude-opus-5} & high & 17--23 Aug \\
Claude Sonnet 5 & \texttt{claude-sonnet-5} & high & 17--20 Aug \\
GPT-5.6 Sol & \texttt{gpt-5.6-sol} & xhigh & 18--28 Aug \\
Claude Opus 5 & \texttt{claude-opus-5} & xhigh & 21--28 Aug \\
\bottomrule
\end{tabular}
\end{table}

\subsection{Statistical power}
For the primary paired comparison, the relevant variation is the spread of
the 27 lineage changes $H_2-H_0$, whose sample standard deviation is 11.92
(standard error 2.29). Assuming this SD for future independent lineages, a
two-sided paired test at $\alpha=.05$ reaches 80\% power for a mean change of
about 6.4 (normal approximation) to 6.7 (noncentral $t$) \rhae{} points at 27
lineages; Table~\ref{tab:power} gives the lineages needed for other effects.
The 3.99 SD of Appendix~\ref{app:fixed-harness-variance} is the run-to-run
spread of one harness's suite score, not the uncertainty of the pooled mean.
Repeats alone may not close this gap: if every snapshot had that execution
SD, with independent errors, the remaining between-lineage variation would
still leave a detectable effect of about 5.9 at 27 lineages with unlimited
repeats per endpoint. This rests on an untested assumption; with an endpoint
SD of 8, unlimited repeats would reduce it to about 2.1. All of these are
planning approximations.

\begin{table}[H]
\caption{Lineages needed for 80\% power (two-sided $\alpha=.05$), assuming
the observed paired-change SD of 11.92. The balanced column rounds the
noncentral-$t$ total up to equal lineages in each of the nine conditions.}
\label{tab:power}
\centering
\small
\begin{tabular}{rrrr}
\toprule
Target gain & Normal approx. & Noncentral $t$ & Balanced (per condition) \\
\midrule
3 & 124 & 126 & 126 (14) \\
5 & 45 & 47 & 54 (6) \\
10 & 12 & 14 & 18 (2) \\
\bottomrule
\end{tabular}
\end{table}

\paragraph{Cost of larger designs.}
The recorded list-price costs (Table~\ref{tab:outer-cost}) indicate what
larger designs would require. A further revision stage for all 27 lineages
would cost about \$5,200 if it cost as much as revision~2 (675 game
evaluations plus coding and qualification). Ten evaluations of every snapshot
instead of one would add about \$145,000 (nine further passes at \$16,130),
and three paired held-out repeats of $H_0$ and $H_2$ for every lineage about
\$28,000. The 126 lineages of Table~\ref{tab:power} would cost about \$85,000
at the observed mean of \$671 per lineage. These extrapolations assume
unchanged per-stage costs and condition mix. The paired repeats and the
crossed evaluation reported in this appendix (810 game evaluations) are
excluded from Table~\ref{tab:outer-cost}.

Table~\ref{tab:stagewise} decomposes the endpoint by stage. The gap is 4.13
after one revision and 5.62 after two; the increase, $+1.49$
$[-1.33,4.58]$, is uncertain, and we do not extrapolate it to further
revisions.

\begin{table}[H]
\caption{Paired stage changes over all 27 lineages. The gap is development
minus held-out change. Intervals are 95\% percentiles of 100,000
whole-lineage resamples under a separate seed (20261108); differences of a
few hundredths from intervals in the main text are Monte Carlo error.
``$+/-/=$'' counts held-out gains, losses, and ties. The two transitions of a
lineage are dependent.}
\label{tab:stagewise}
\centering
\scriptsize
\setlength{\tabcolsep}{4pt}
\begin{tabular}{lrrrc}
\toprule
Transition & Development change & Held-out change & Gap & $+/-/=$ \\
\midrule
$H_0\!\to\!H_1$ & $+5.46$ $[0.73,10.55]$ & $+1.33$ $[-2.47,4.96]$ &
$+4.13$ $[0.60,7.66]$ & 15/11/1 \\
$H_1\!\to\!H_2$ & $+2.96$ $[-0.30,6.31]$ & $+1.47$ $[-0.84,4.13]$ &
$+1.49$ $[-1.33,4.58]$ & 12/15/0 \\
$H_0\!\to\!H_2$ & $+8.42$ $[2.81,14.76]$ & $+2.80$ $[-1.41,7.41]$ &
$+5.62$ $[1.81,9.75]$ & 15/12/0 \\
\bottomrule
\end{tabular}
\end{table}

\subsection{Selection-rule provenance and sensitivity}
The development-based rule of Table~\ref{tab:checkpoint-selection} was
specified post hoc. The 24 lineages of the eight conditions other than GPT-6
Astra ran 17--28 August 2026, and their held-out scores were tabulated on
31~August. The rule first appears in our version history on 3~September
(local time, UTC+2), already labeled a post hoc diagnostic, in a commit that
descends from the one containing the held-out table. No analysis plan was
preregistered: the campaign manifest of 17~August fixes replicate counts,
prompts, and the replacement policy, but no endpoint or selection rule. Our code and notes contain no alternative development-based rule, and
the rule has not changed since. Only the three GPT-6 Astra lineages, run
13--15 September, postdate it; on them it gains $+0.08$ over deploying $H_2$,
whose mean is 94.37, so they provide no validation.

Table~\ref{tab:selection-loo} recomputes the selection after omitting one
development game at a time, with all candidates fixed. Averaging over all
subsets of a given size within each lineage, gains over $H_0$ are $+4.08$
$[0.40,8.19]$, $+3.77$ $[0.12,7.81]$, $+3.49$ $[-0.03,7.41]$, and $+2.64$
$[-0.16,5.75]$ for four, three, two, and one selection games; gains over $H_2$
are $+1.28$ $[0.04,2.95]$, $+0.97$ $[-0.31,2.54]$, $+0.69$ $[-0.70,2.22]$, and
$-0.17$ $[-2.13,1.72]$. Evidence for a positive gain over $H_0$ is thus more
stable than for an advantage over always deploying $H_2$. All candidates were
revised using feedback from all five games, so an omitted game remains in the
adaptation history: this checks the selector's sensitivity, not adaptation,
and is not a prospective validation.

\begin{table}[H]
\caption{Leave-one-development-game-out selection. Each row selects among the
fixed $H_0$, $H_1$, and $H_2$ by mean development \rhae{} on the remaining
games, breaking ties toward the earlier snapshot. ``$+/-/=$'' counts held-out
improvement, regression, and no change relative to $H_0$. Intervals use
100,000 whole-lineage resamples under a separate seed (20261108), so the
all-five row differs from Table~\ref{tab:checkpoint-selection} by Monte Carlo
error.}
\label{tab:selection-loo}
\centering
\scriptsize
\setlength{\tabcolsep}{4pt}
\begin{tabular}{lcrcr}
\toprule
Omitted game & $H_0/H_1/H_2$ & Gain vs.\ $H_0$ & $+/-/=$ & Gain vs.\ $H_2$ \\
\midrule
None (all five) & 9/6/12 & $+4.53$ $[0.80,8.71]$ & 13/5/9 & $+1.72$ $[0.38,3.68]$ \\
\texttt{tr87} & 8/6/13 & $+3.95$ $[-0.19,8.40]$ & 13/6/8 & $+1.15$ $[0.36,2.14]$ \\
\texttt{vc33} & 8/6/13 & $+4.13$ $[0.41,8.36]$ & 13/6/8 & $+1.33$ $[-0.21,3.39]$ \\
\texttt{r11l} & 12/7/8 & $+4.09$ $[0.35,8.33]$ & 10/5/12 & $+1.29$ $[-0.51,3.43]$ \\
\texttt{bp35} & 9/7/11 & $+4.09$ $[0.68,7.79]$ & 13/5/9 & $+1.28$ $[-0.49,3.41]$ \\
\texttt{ft09} & 7/7/13 & $+4.16$ $[0.16,8.52]$ & 14/6/7 & $+1.36$ $[-0.10,3.38]$ \\
\bottomrule
\end{tabular}
\end{table}

\subsection{Floor conditions and builder/player coupling}
Table~\ref{tab:floor} removes near-floor conditions while keeping all three
lineages of every retained condition. The thresholds use only condition-mean
$H_0$ development \rhae{}; they are post hoc baseline subsets that change the
target population rather than correct the primary estimate.

\begin{table}[H]
\caption{Floor-condition sensitivity. Thresholds apply to condition-mean
$H_0$ development \rhae{}: $>1$ drops Sonnet, $>5$ also Luna, and $>10$ also
Terra, which coincides with excluding all three. The gap is development minus
held-out change. Intervals are 95\% percentiles of 100,000 whole-lineage
resamples (seed 20261108).}
\label{tab:floor}
\centering
\scriptsize
\setlength{\tabcolsep}{4pt}
\begin{tabular}{lrrrrc}
\toprule
Cohort & Lineages & Held-out $H_2-H_0$ & Dev.\ $H_2-H_0$ & Gap & Gains/losses \\
\midrule
All conditions & 27 & $+2.80$ $[-1.41,7.41]$ & $+8.42$ & $+5.62$ $[1.81,9.75]$ & 15/12 \\
Excluding Terra, Luna, Sonnet & 18 & $+4.82$ $[-1.44,11.38]$ & $+13.34$ & $+8.52$ $[3.16,14.11]$ & 12/6 \\
$H_0$ development $>1$ & 24 & $+3.12$ $[-1.66,8.29]$ & $+9.44$ & $+6.32$ $[2.09,10.83]$ & 13/11 \\
$H_0$ development $>5$ & 21 & $+3.64$ $[-1.79,9.41]$ & $+10.85$ & $+7.22$ $[2.45,12.22]$ & 12/9 \\
$H_0$ development $>10$ & 18 & $+4.82$ $[-1.44,11.38]$ & $+13.34$ & $+8.52$ $[3.16,14.11]$ & 12/6 \\
\bottomrule
\end{tabular}
\end{table}

Within a lineage the runtime model is fixed, so $H_0\!\to\!H_2$ changes
reflect revision plus execution noise. The near-floor harnesses run: all 675
Terra, Luna, and Sonnet games take charged actions and pass controller
checks, 4.3\% end in a nonzero harness exit not caused by an action cap
(6.7\% in the other six conditions), and Terra and Luna complete at least one
level in 48\% and 44\% of their games (Sonnet 10\%). With the same Terra
player, held-out level completions fall from 64 at $H_0$ to 28 at $H_1$ and
22 at $H_2$, lower in all three lineages; weak play also thins the feedback,
as five of Sonnet's six development bundles contain no completed level. Floor compression
shrinks both terms of the gap; the three conditions together contribute
$-0.06$ of its 5.62 points.

\paragraph{Crossed builder/player evaluation.}
To separate construction from gameplay, we re-evaluated frozen $H_0$ and $H_2$
snapshots with a different runtime model, changing only the controller-written
model profile; generated code is byte-identical. All three Sol-high lineages
were played by Terra-high and by Luna-high, and all three Sonnet-high lineages
by Sol-high, each once on the 20 held-out games (36 suites including the
prespecified development suites). Table~\ref{tab:crossplay} compares each
crossed cell with the same harness played by its builder (player effect) and
with the player's own harness (builder effect).
For Terra and Luna, gameplay limits the score: Sol-built harnesses
lose 19--26 points when they play them, yet gain them only 1--6 points over
their own harnesses. For Sonnet, the harness does: its harnesses run without
errors, but Sol gains only 1--2 points when playing them, whereas Sol's own
harnesses score about 25 points higher. The reverse cells (a Terra- or Luna-built harness played by
Sol, a Sol-built harness played by Sonnet) were not run, so these models'
other role is not assessed separately. Weak players also compress revision effects: under Terra or
Luna, the Sol lineages' $H_2-H_0$ changes ($+33.18$, $-15.18$, $-10.70$ with
Sol) lie between $-5.51$ and $+8.20$. With one evaluation per cell, these
comparisons are descriptive.

\begin{table}[H]
\caption{Crossed builder/player evaluation on the 20 held-out games. Entries
are mean differences in \rhae{} over three lineages, with ranges. Player
effect: crossed cell minus the same harness played by its builder. Builder
effect: crossed cell minus the player's own harness of the same replicate.}
\label{tab:crossplay}
\centering
\scriptsize
\setlength{\tabcolsep}{3pt}
\resizebox{\linewidth}{!}{%
\begin{tabular}{llrrrrr}
\toprule
Harness & Player & \multicolumn{2}{c}{Player effect} & \multicolumn{2}{c}{Builder effect} & Revision $H_2-H_0$ \\
\cmidrule(lr){3-4}\cmidrule(lr){5-6}
 & & $H_0$ & $H_2$ & $H_0$ & $H_2$ & crossed \\
\midrule
Sol high & Terra high & $-18.80$ $[-21.37,-16.71]$ & $-21.72$ $[-41.69,-10.34]$ & $+3.18$ $[-1.10,8.50]$ & $+6.17$ $[1.37,11.38]$ & $-0.49$ $[-5.51,8.20]$ \\
Sol high & Luna high & $-23.52$ $[-27.92,-17.72]$ & $-26.37$ $[-50.23,-12.16]$ & $+1.18$ $[1.14,1.25]$ & $+1.25$ $[-0.17,2.48]$ & $-0.41$ $[-2.41,0.67]$ \\
Sonnet high & Sol high & $+0.79$ $[0.02,2.22]$ & $+2.05$ $[1.29,2.48]$ & $-25.27$ $[-27.79,-20.24]$ & $-26.18$ $[-52.16,-9.45]$ & $+1.53$ $[0.16,3.15]$ \\
\bottomrule
\end{tabular}}
\end{table}

\subsection{Paired repeated evaluations}
\label{app:paired-repeats}

To test whether individual endpoint classifications survive execution noise,
we re-evaluate $H_0$ and $H_2$ of three lineages three times each on all 20
held-out games (360 game evaluations). The lineages were fixed from
historical outcomes before any new inference. The near-zero case has the
smallest absolute change among lineages with both endpoint means in
$[5,90]$ and an absolute change below 5 (Fable-high replicate~1,
$16.05\to13.77$); the gain and loss cases are the largest gain and largest
loss among the Sol-high lineages (replicate~1, $20.38\to53.56$; replicate~2,
$27.92\to12.74$). For the first two lineages, $H_0$ and $H_2$ run
concurrently in each of three sequential blocks; the loss lineage's six
suites run concurrently afterwards, so its three pairs are not separated in
time. Concurrency exceeds that of the original campaign. Harness, controller,
model-profile, provider-configuration, and environment fingerprints are verified
against the original evaluations, and no construction, revision, repair,
hosted replay, or selection occurs.

Because provider-side changes since August cannot be excluded, fresh $H_0$ is
compared with fresh $H_2$, not with the original scores. Each pair's
difference averages the 20 paired game differences; we report the mean of
the three differences with a paired $t$ interval (df${}=2$), call a direction
supported only if that interval excludes zero, and use the $\pm5$-point band
as a second classification. If both endpoints have an SD near 4 and are
independent, the interval half-width is about 14 points, and because the gain and loss cases are
extremes, some shrinkage toward zero is expected. Games whose failure carries
an explicit provider-infrastructure classification are resumed by the
controller's standard mechanism, as in the original campaign; unexplained
incomplete attempts and blocked validations stop the campaign. This policy
was adopted after one such failure in the first block, before that suite had
a computed score; the means of the other three first-block suites had
already been computed.

All 18 suites completed. Fifteen of the 360 games (4.2\%) hit an explicit
provider-infrastructure failure (hard deadline or HTTP 504) and completed
cleanly after one resume.
Table~\ref{tab:paired-repeats} reports the results. Both large observed changes reproduce: the gain
lineage improves in all three fresh pairs ($+30.82$, interval $[9.97,51.67]$)
and the loss lineage declines in all three ($-16.57$, $[-28.16,-4.99]$). The
loss interval's upper end lies just inside the $\pm5$ band, so its direction
is supported while its practical class is unresolved under the preset rule.
The near-zero lineage remains unresolved ($-1.15$, $[-9.64,7.33]$). No case
supports a strict score--cost classification: every cost-per-game interval
includes zero.

\begin{table}[H]
\caption{Paired repeated held-out evaluations. Each fresh difference is
$H_2-H_0$ in mean held-out \rhae{} over the same 20 games; intervals are paired
$t$ with df${}=2$. Direction is supported if the interval excludes zero; the
$\pm5$ class requires the whole interval beyond the band.}
\label{tab:paired-repeats}
\centering
\scriptsize
\setlength{\tabcolsep}{3.5pt}
\begin{tabular}{lrrrrrll}
\toprule
Lineage & Original $H_0\!\to\!H_2$ & Pair 1 & Pair 2 & Pair 3 & Mean $[95\%]$ & Direction & $\pm5$ class \\
\midrule
Fable high, rep.~1 & $16.05\to13.77$ & $-4.41$ & $+2.40$ & $-1.45$ & $-1.15$ $[-9.64,7.33]$ & unresolved & unresolved \\
Sol high, rep.~1 & $20.38\to53.56$ & $+35.05$ & $+36.25$ & $+21.15$ & $+30.82$ $[9.97,51.67]$ & gain & gain \\
Sol high, rep.~2 & $27.92\to12.74$ & $-21.96$ & $-14.02$ & $-13.74$ & $-16.57$ $[-28.16,-4.99]$ & loss & unresolved \\
\bottomrule
\end{tabular}
\end{table}

\subsection{Rule-based check of interpreted codes}
The two codes interpreted in Section~\ref{sec:revision-behavior},
\textsc{Scope} and binary \textsc{Adapt} (further result-driven action versus
handoff), are defined by file counts, line counts, and event order. We
re-derived them with a deterministic script that uses no language model;
its rules and implementation hash were frozen before any per-episode label,
rationale, or outcome was opened. It computes \textsc{Scope} from the released
parent and child snapshots and \textsc{Adapt} from the packets' chronological
tool events, the evidence the coders saw (Table~\ref{tab:coding-validation}).
\textsc{Scope} agrees closely and reproduces the published contrast; its three
disagreements depend on how the codebook's line-count denominator or
subsystem granularity is read. \textsc{Adapt} agrees less ($\kappa=.64$, below the .67
inclusion threshold), and its contrast does not hold: the published negative
mean rests on seven episodes that only the consensus codes as reactive (mean
change $-5.38$), while the 11 episodes that both code as reactive average
$+2.18$. Under script labels, reactive and handoff episodes also start from
similar parents (24.65 versus 25.36 held-out \rhae{}). The \textsc{Adapt}
contrast therefore depends on how post-result adaptation is operationalized.
The script is a second operationalization with its own errors, not a human
validation.

\begin{table}[H]
\caption{Deterministic re-derivation versus the LLM consensus over 54
episodes. Agreement is exact; $\kappa$ is quadratic-weighted for
\textsc{Scope} and unweighted for binary \textsc{Adapt}. Contrasts follow
Table~\ref{tab:process-results} (scoped minus broad; further action minus
handoff) with the same resampling procedure and seeds. Script variants were
fixed before the comparison.}
\label{tab:coding-validation}
\centering
\scriptsize
\setlength{\tabcolsep}{4pt}
\begin{tabular}{llrrr}
\toprule
Code & Labels & Agreement & $\kappa$ & Contrast [95\% interval] \\
\midrule
\textsc{Scope} & LLM consensus (published) & -- & -- & $-3.94$ $[-9.09,1.22]$ \\
 & Script, role-group subsystems & 51/54 & .90 & $-3.77$ $[-9.35,1.62]$ \\
 & Script, module subsystems & 52/54 & .91 & $-3.60$ $[-9.15,1.74]$ \\
\addlinespace[2pt]
\textsc{Adapt} & LLM consensus (published) & -- & -- & $-3.24$ $[-8.54,2.38]$ \\
 & Script, failure-triggered & 46/54 & .64 & $+0.48$ $[-5.09,7.05]$ \\
 & Script, any post-edit result & 43/54 & .55 & $-2.39$ $[-6.25,2.72]$ \\
 & Script, structural smoke repair & 48/54 & .74 & $-0.17$ $[-5.23,5.70]$ \\
\bottomrule
\end{tabular}
\end{table}

\subsection{Recovery sensitivity}
All 2,025 selected scores, recomputed from the original local scorecards with
the released scorer, match. Of the 105 additional attempts, 49 carry an
explicit provider-infrastructure reason (42 HTTP 504 responses, six hard
deadlines at 1,830 seconds, and one upstream timeout at 3,540 seconds), and
11 are blocked by validation checks (five model-response replay divergences,
three game-replay divergences, and three missing local scorecards). The
other 45 are incomplete: each retains one unresolved model request and no
exit status or outcome reason, so we leave their cause unresolved. None is
classified as a model-software failure; harness error exits occur within
clean selected attempts and are scored as outcomes. The 45 incomplete
attempts occupy 38 slots, 11 of which also hold one of the 17 selected fresh
restarts, so 44 slots in eight lineages are affected
(Table~\ref{tab:recovery-dist}); all fresh restarts are on held-out games.

\begin{table}[H]
\caption{Recovery events by snapshot and split. Incomplete attempts lack an
outcome reason; parentheses give their distinct slots. Fresh restarts are
selected attempts without a replay parent. The union counts each affected
slot once.}
\label{tab:recovery-dist}
\centering
\small
\begin{tabular}{llrrrr}
\toprule
Snapshot & Split & Incomplete attempts (slots) & Fresh restarts & Overlap & Union \\
\midrule
$H_0$ & Development & 7 (4) & 0 & 0 & 4 \\
$H_0$ & Held-out & 9 (7) & 6 & 3 & 10 \\
$H_1$ & Development & 0 (0) & 0 & 0 & 0 \\
$H_1$ & Held-out & 0 (0) & 1 & 0 & 1 \\
$H_2$ & Development & 3 (3) & 0 & 0 & 3 \\
$H_2$ & Held-out & 26 (24) & 10 & 8 & 26 \\
\midrule
All & & 45 (38) & 17 & 11 & 44 \\
\bottomrule
\end{tabular}
\end{table}

Table~\ref{tab:recovery-sens} removes affected games at both endpoints and
averages paired changes with equal lineage weights. Under the union
exclusion, the held-out gain moves from 2.80 to 2.56 and the gap from 5.62 to
4.21 with unchanged sign counts, but the gap interval, $[-0.05,8.87]$, then
includes zero. This exclusion retains all 27 lineages but leaves as few as 11
held-out games and, for one lineage, a single development game. Excluding the
seven lineages with an affected endpoint slot entirely gives a gap of 5.20
but a held-out gain of only 1.02; all seven had positive original changes, so
this exclusion is not missing at random. Replacing each affected endpoint
score in the union by any value in $[0,100]$ bounds the held-out gain to
$[-0.21,6.46]$, the gap to $[-1.38,10.47]$, and losses to 12--15; these are
deterministic bounds, not confidence intervals. The gap's point estimate thus
stays positive under every exclusion, while its interval, and its sign under
worst-case bounds, depend on how recovery is treated.

\begin{table}[H]
\caption{Paired exclusion sensitivity. Rows remove the listed endpoint slots
at both $H_0$ and $H_2$ of the same lineage and game; the last row removes
whole lineages. Pairs count held-out/development game pairs. Intervals, where
computed, are 95\% percentiles of 20,000 whole-lineage resamples (seed
20261108), conditional on each row's retained games.}
\label{tab:recovery-sens}
\centering
\scriptsize
\setlength{\tabcolsep}{3.5pt}
\begin{tabular}{lrrrrrc}
\toprule
Excluded & Lineages & Pairs & Held-out gain & Dev.\ gain & Gap & Gains/losses \\
\midrule
None & 27 & 540/135 & $2.80$ $[-1.39,7.45]$ & 8.42 & $5.62$ $[1.80,9.74]$ & 15/12 \\
Incomplete-attempt slots & 27 & 514/128 & 2.54 & 6.77 & 4.23 & 15/12 \\
Fresh-restart slots & 27 & 526/135 & 2.86 & 8.42 & 5.56 & 14/13 \\
Union of both & 27 & 509/128 & $2.56$ $[-1.64,7.20]$ & 6.77 & $4.21$ $[-0.05,8.87]$ & 15/12 \\
Any multiple-attempt endpoint slot & 27 & 501/125 & $3.02$ $[-1.29,7.82]$ & 6.65 & $3.62$ $[-0.76,8.42]$ & 15/12 \\
Seven lineages with a union slot & 20 & 400/100 & $1.02$ $[-4.24,7.01]$ & 6.21 & $5.20$ $[0.59,10.27]$ & 8/12 \\
\bottomrule
\end{tabular}
\end{table}

\end{document}